\documentclass[fleqn,10pt]{wlscirep}
\usepackage[utf8]{inputenc}
\usepackage[T1]{fontenc}

\usepackage{tabularx}
\usepackage{multirow}
\usepackage{booktabs, tabu}
\usepackage{algorithm}
\usepackage[noend]{algpseudocode}
\usepackage{lipsum,graphicx}
\usepackage{subfigure}
\usepackage{algorithm}
\usepackage[noend]{algpseudocode}

\title{A Whole Slide Image Grading Benchmark and Tissue Classification for Cervical Cancer Precursor Lesions with Inter-Observer Variability}

\author[1,5]{Abdulkadir Albayrak}
\author[2]{Asli Unlu}
\author[3]{Nurullah Calik}
\author[4]{Abdulkerim Capar}
\author[1,5]{Gokhan Bilgin}
\author[4]{Behcet Ugur Toreyin}
\author[2]{Bahar Muezzinoglu}
\author[2]{Ilknur Turkmen}
\author[4,*]{Lutfiye Durak-Ata}

\affil[1]{Yildiz Technical University, Dept. of Computer Engineering, Istanbul, 34220, TURKEY}
\affil[2]{Istanbul Medipol University Hospital, Dept. of Pathology, Istanbul, 34214, TURKEY}
\affil[3]{Yildiz Technical University, Dept. of Electronics and Telecommunications Engineering, Istanbul, 34220, TURKEY}
\affil[4]{Istanbul Technical University, Informatics Institute, Istanbul, 34469, TURKEY}
\affil[5]{Yildiz Technical University, Signal and Image Processing Laboratory (SIMPLAB), Istanbul, 34220, TURKEY}
\affil[*]{Corresponding Author: durakata@itu.edu.tr}


\begin{abstract}
The cervical cancer developing from the precancerous lesions caused by the Human Papilloma Virus (HPV) has been one of the preventable cancers with the help of periodic screening. There are two types of grading conventions widely accepted among pathologists. On the other hand, inter-observer variability is an important issue for final diagnosis. In this paper, a whole-slide image grading benchmark for cervical cancer precursor lesions is introduced. The papillae of the cervical epithelium and overlapping cell problems are handled and a tissue classification method with a novel morphological feature exploiting the relative orientation between the BM and the major axis of all nuclei is developed and its performance is evaluated. Besides, the inter-observer variability is also revealed by a thorough comparison among pathologists’ decisions, as well as, the final diagnosis.
\end{abstract}
\begin{document}

\flushbottom
\maketitle
%
%
\thispagestyle{empty}


\section*{Introduction}

\label{sec:introduction} 
Cervical cancer is one of the most commonly seen cancer type in the world and the 4th most common cause of death, which develops from precursor lesions \cite{torre2015global}. Studies indicate that, almost all cervical cancer cases develop by the effect of Human Papilloma Virus (HPV), which reaches epithelial basal layer cells with the help of micro-injuries in the cervical epithelium. Carcinogenic effect of the virus occurs when HPV's genome integrates with the cell genome \cite{stoler2000human,van2018hpv,zur2009papillomaviruses}. This effect, which requires a certain period of time, appears as morphological changes in the cervical epithelium. These precancerous lesions characterized by dysplastic changes are called squamous intraepithelial lesions (SIL).

 Impact of HPV on the cervical epithelium varies throughout the life cycle of the virus, which in turn results in different morphological changes. Early diagnosis can be made possible by the analysis of these morphological structures~\cite{torre2015global,stoler2000advances}. After being infected by HPV, basal cells proliferate and the epithelium loses its maturation. As well as the loss of maturation which results in polarity loss in the epithelium, cells show nuclear enlargement, nuclear irregularity, and hyperchromasia. Depending on the proliferation process, the number of mitoses also increases. The effect of viral proteins on the cyto-skeleton reveals halo cells with characteristic perinuclear halo named “koilocytes” (it means "hollow" in Greek). These dysplastic changes are graded according to whether they are seen among the lower, middle and upper part of the epithelium. They are reported according to Cervical Intraepithelial Neoplasia (CIN) 1-3 in the CIN-based grading and LSIL (low-grade squamous intraepithelial lesion) and HSIL (high-grade squamous intraepithelial lesion) in the SIL-based grading \cite{cox2013historical,darragh2012lower,mitra2015cervical}. Currently, the use of SIL-based grading is recommended, yet the CIN-based grading is also used. 

Pathologic diagnosis of cervical biopsies varies depending on biopsy ingestion or artifacts due to laboratory steps, and pathological interpretation. Due to the spread of women health screening programs, the diagnosis of cervical biopsies is frequently encountered and this diagnosis variability has become a more important problem. Cervical biopsy interpretation has inter- or intra-observer variability, which means that a biopsy may have different diagnoses by different pathologists or by the same pathologist at different times, and it is accepted to an extent in the literature \cite{stoler2001interobserver,mccluggage1996interobserver,mccluggage1998inter}.  Studies have been made to overcome this problem with classification systems suitable for the nature of HPV or with the help of immunohistochemical techniques \cite{doorbar2007papillomavirus,stoler2000human,darragh2012lower}.

Increasing role of the information technology (IT) on the area of medicine has a positive impact on the pathology. Digital pathology that includes diagnosis, education, consultation, archiving, and also morphometric evaluation tools \cite{al2012digital,madabhushi2016image}.  Studies about morphometric analysis are available for different tissues and systems \cite{he2012histology}, as well as for cervical lesions \cite{de2013fusion,guo2016nuclei,naghdy2012computer,wang2009assisted} in the literature. De et al. \cite{de2013fusion} studied image analysis methods on 62 digital images of cervical epithelial with Normal, CIN1, CIN2 and CIN3 labeled lesions.  The cervical regions are manually marked by the pathologist on selected epithelial images, and these regions are divided into vertical segments by calculating the medial axis. The obtained epithelial segments are examined in terms of structural, geometric, and profile-based properties. Contrast, energy levels, pixel correlation values, and neighborhood features of the pixels within the vertical segment are studied as the structural features. Geometric features include the distances between nuclei centers and Delaunay triangulation. 
In the profile-based feature extraction, correlation values and the brightness values of all pixels of each row of vertical segment is calculated. Linear Discriminant Analysis (LDA) and Support Vector Machines (SVM) are utilized to classify feature vectors of vertical segments. First, each of the vertical segments is classified individually, then these decisions are fused to obtain a whole epithelium classification result. The effect of individual decisions of vertical segments on the whole epithelium classification result is also examined. One-to-one correspondence between the system result and pathological diagnosis is named as "Exact Class Label" (1st approach), only one class difference between system result and pathological diagnosis is named as "Windowed Class Label" (2nd approach) and bigger differences between system result and pathological diagnosis is named as Normal versus CIN (3rd approach). Different classification performances are calculated using different approaches and features.  Using all the structural, geometric and profile-based features, recognition rate of 62.3\% on vertical segments and 39.3\% on whole epithelium is reached.

Guo et al. \cite{guo2016nuclei} tried to develop enhanced image analysis methods on the cervix image data set that was formed in De et al. \cite{de2013fusion}. They have increased the success of classification by adding structural features of the nucleus and cytoplasm in addition to the features extracted from similar vertical segments as in De et al. \cite{de2013fusion}. These features consist of nucleus, cytoplasm, and acellular areas and ratios, color scale (red, green, blue) brightness values, numbers of triangles obtained by Delaunay triangulation at upper, middle, and lower epithelium regions. The features are classified by the classification methods of the previous study. The name "Windowed Class Label" used in the second approach in the previous study is changed to "Of-By-One Class Label". In this study, as well as using the same data set as De et al., they made a difference of examination by 2 different pathologists. The diagnostic success rates of the extracted features are determined by the Attribute Information Gain Ratio (AIGR) algorithm. They evaluated the success of features according to two different classification approaches. As a result of adding structural features of cervical regions, they have increased their classification success up to 82-88.5\%.

\begin{figure}[!ht]
\begin{center}
\begin{tabular}{c}
\includegraphics[width=\textwidth]{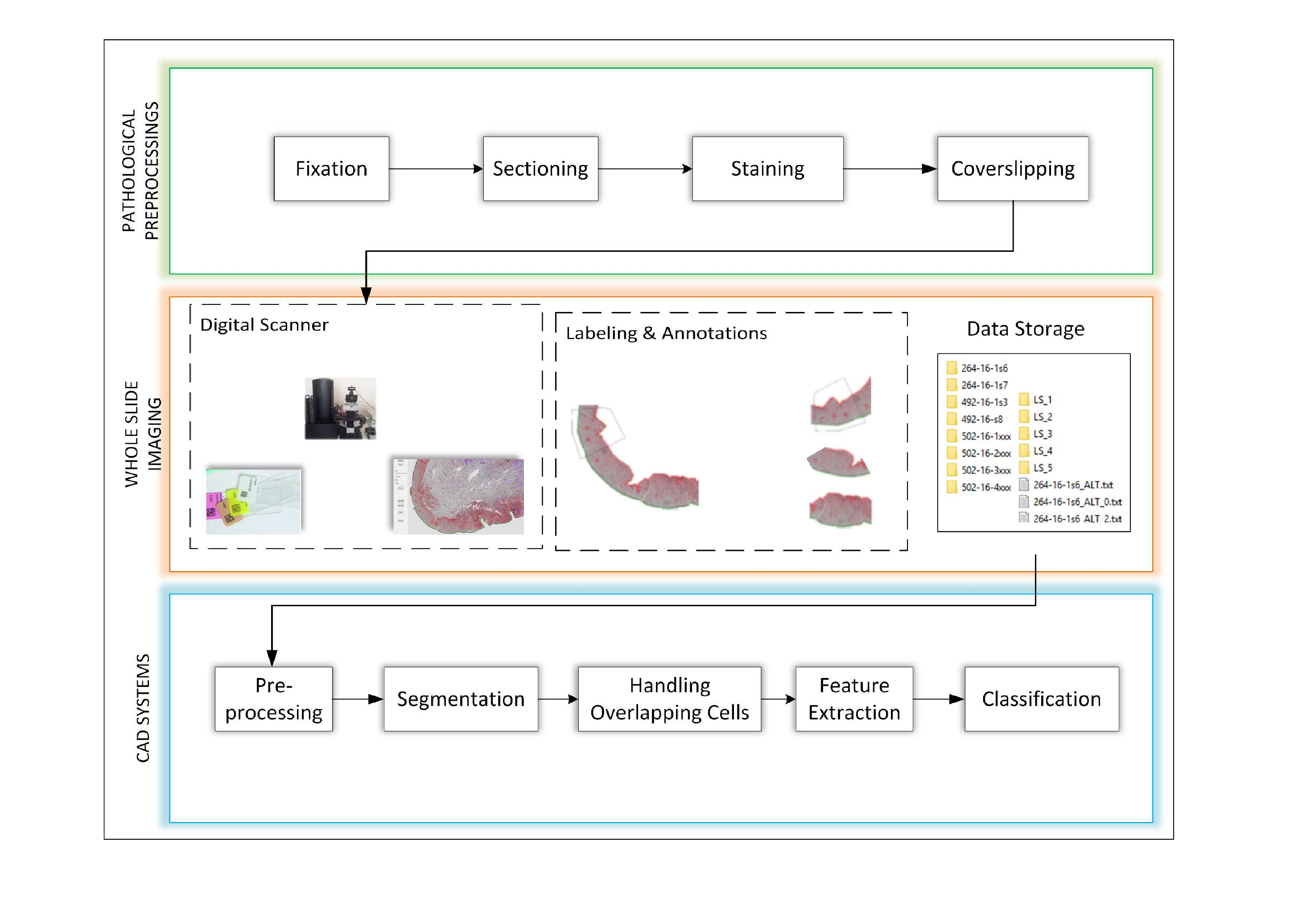}
\end{tabular}
\end{center}
\caption[]{Processing steps followed in the proposed study.  The first step describes the pathological preprocess which is handled in the pathology laboratory. Whole slide scanning and the filing process mentioned in the second row is done by the medical researcher and the computer scientist in collaboration. CADAS framework developed for grading the cervical cancer pre-cursor lesion is mentioned in the third row.}
\label{fig:processingSteps}
\end{figure} 

Wang et al. studied morphometric analysis methods on 31 digital images of cervical biopsies \cite{wang2009assisted}. Their study consists of two steps as the automated segmentation of squamous epithelium and the CIN classification. In the first step, the epithelium is segmented using the difference of the visual properties of five different regions consisting of squamous epithelium, columnar epithelium, stroma, background and erythrocytes. The medial axis is drawn parallel to the basal and upper membrane borders after the epithelial region is segmented. Square windows with $250\times250$ pixel dimensions are created on normal lines of medial axis. The features of the nucleus average area and number, the average area of the triangles obtained by Delaunay triangulation, and the average edge length are analyzed within each window. Obtained feature vectors are fed to different classification methods, and have reached accuracy rates ranging from 60\% to 95\%.

Keenan et al. proposed a study to analyze 230 digital cervix images consisting of normal, koilocytosis, CIN1, CIN2, and CIN3 labeled lesions \cite{keenan2000automated}. The features of the nucleus area, the nucleus cytoplasm ratio, the ratio of nucleus area to cytoplasm, and the edges / areas of the Delaunay triangles are analyzed. The Kappa value for the observer difference between the two pathologists involved in the study is 0.415. The classification performance of the system in distinguishing normal and CIN lesions is 98.7\%. The overall success rate is 62.3\% where the worst performance is achieved on CIN2 labeled patterns.

Nagdhy et al. proposed a study to classify a total of 475 cervical biopsies with normal, CIN1, CIN2, CIN3, and invasive carcinoma using three different methods \cite{naghdy2012computer}. The nucleus area, core cytoplasm ratio, core boundary irregularity, and areas of Delaunay triangles are analyzed. They reached up to 97\% with respect to specificity and 100\% with respect to sensitivity using different methods including Gabor-based texture descriptor, GLCM (Gray-level co-occurrence matrix) texture descriptor, and pre-trained convolutional neural network.

In this study, morphometric analysis methods for the cervical SIL diagnosis is investigated on a new digital cervical image data set. The numerical values of the morphological features used by the pathologists in the diagnosis are extracted. The statistical significance of their contribution to diagnosis is examined. Within the scope of the study, a Computer Aided Diagnostic Auxiliary System (CADAS) is developed and its performance is evaluated.

Contributions of this study are as follows:

\begin{itemize}
    \item A new whole slide image grading benchmark for grading of cervical dysplasias is created and introduced to histopathological image analysis community.
    \item Images obtained from the data set are labeled by two pathologists to mention the inter-observer variability in cervical dysplasia grading.
    \item Pathologists diagnosed each image patch stained with hematoxylin $\epsilon$ eosin (H$\epsilon$E) in the data set independently. In the likely case of inconsistent diagnoses, the image patches that are stained with p16 and Ki67 immunhistochemical dyes are analyzed to decide a final diagnosis.
    \item A morphometric analysis method for cervical SIL diagnosis is proposed.
    \item The presence of papillaries in the dataset that leads to tangential sections is one of the important parameters that pathologists give account for when diagnosing.
\end{itemize}

\section{Materials and Methods}
\label{sec:methods}    

This  study  is  conducted  by  a  group  of scientists  and  medical  researchers. Cervical  tissue  slide samples with diagnosis results are collected in the pathology laboratory of Istanbul Medipol University (IMU) Hospital, Istanbul, Turkey. Fig. \ref{fig:processingSteps} shows the processing steps followed in this study for the proposed CADAS to grade cervical cancer precursor lesions. 

\subsection{Data Collection and Image Acquisition} 
\label{imageAcquisition}

\begin{figure*}[!ht]
  \centering
  \subfigure[]{
    \includegraphics[width=2.75cm,height=3cm]{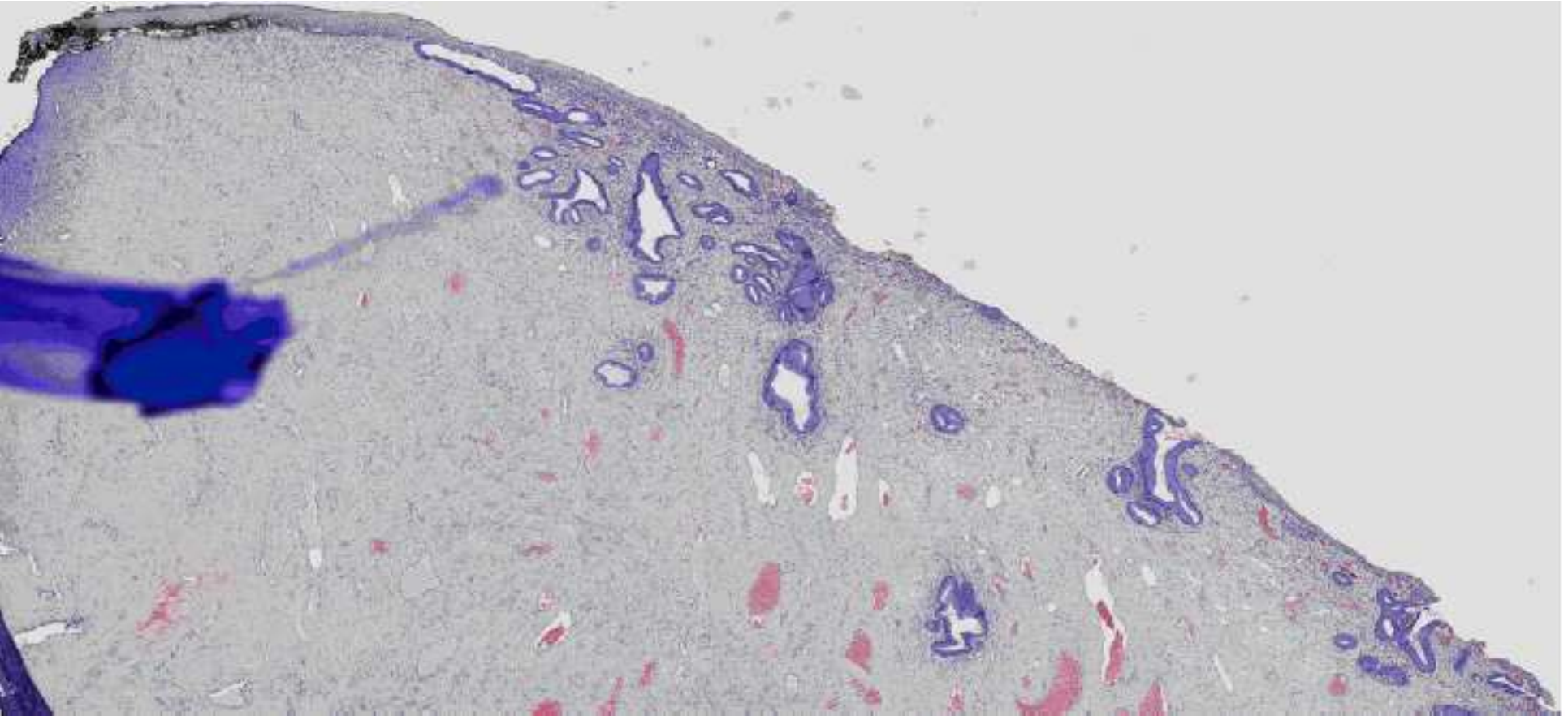}}
  \subfigure[]{
    \includegraphics[width=2.75cm,height=3cm]{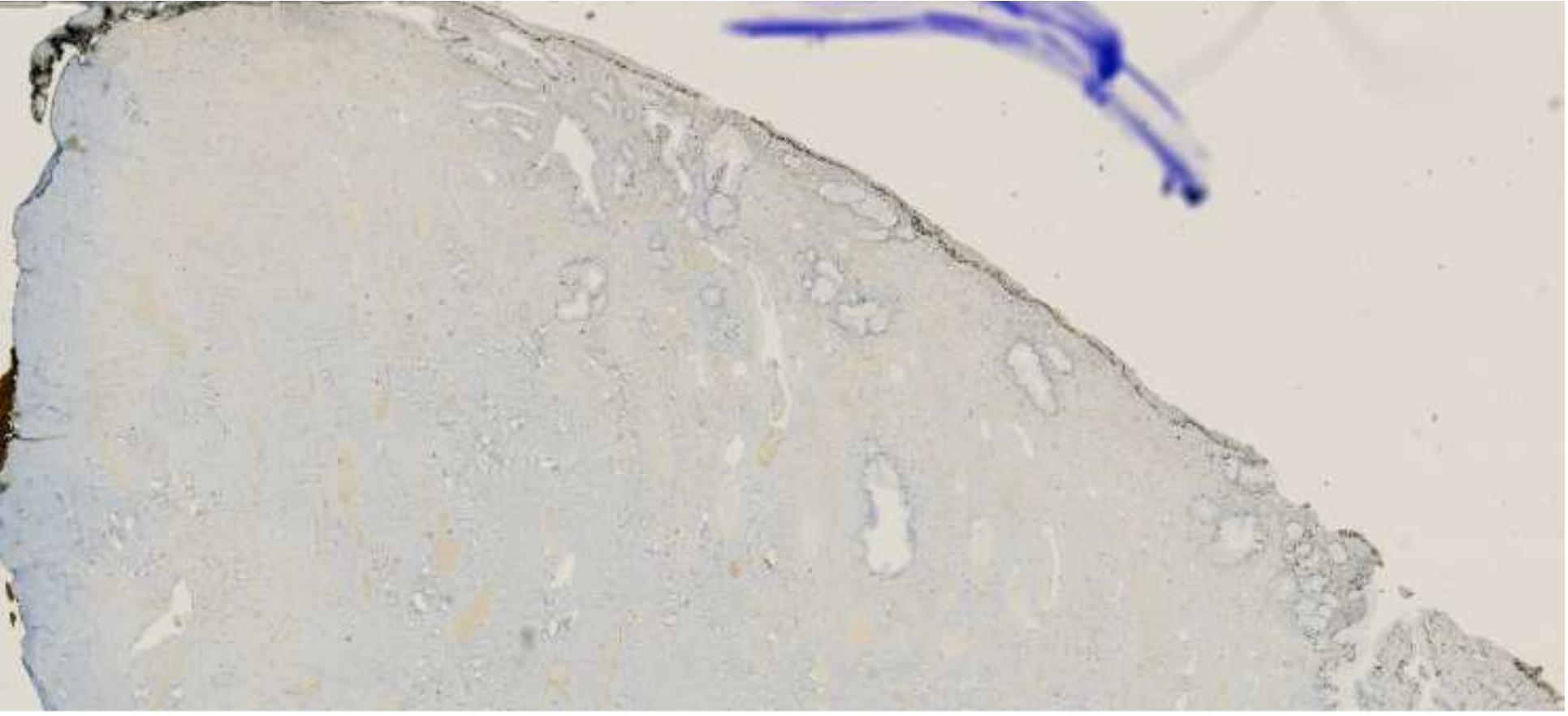}}
  \subfigure[]{
    \includegraphics[width=2.75cm,height=3cm]{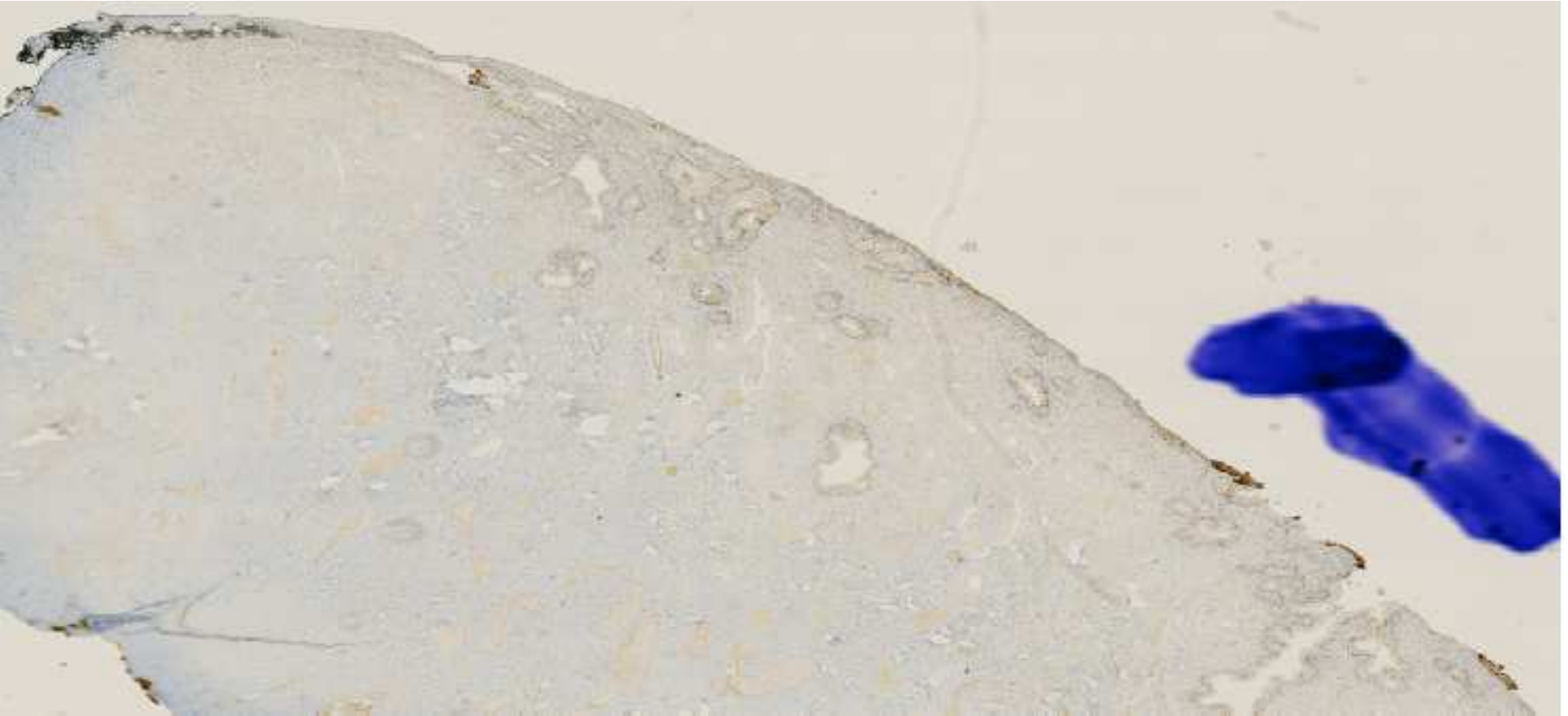}} \\
    \subfigure[]{
    \includegraphics[width=2.75cm,height=3cm]{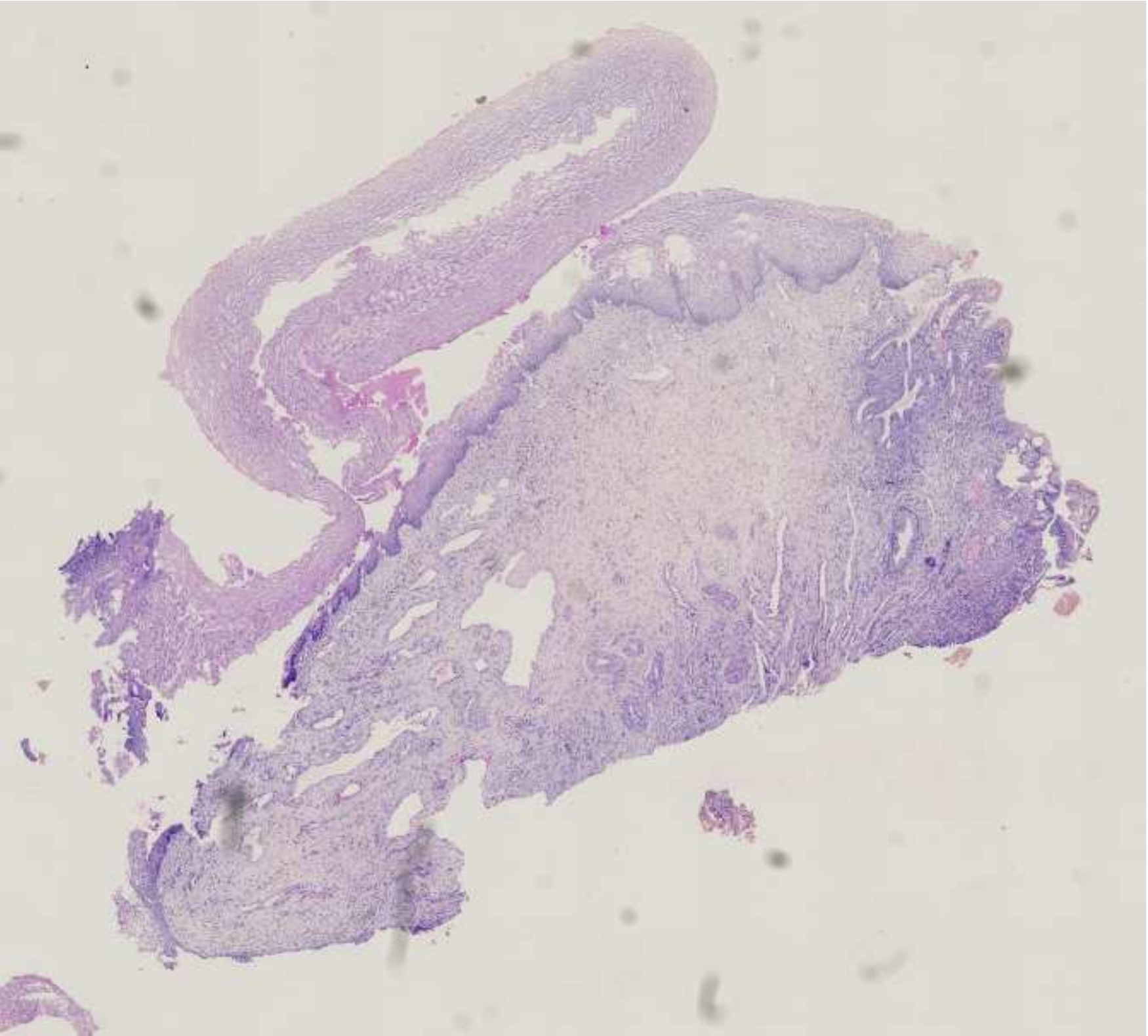}}
  \subfigure[]{
    \includegraphics[width=2.75cm,height=3cm]{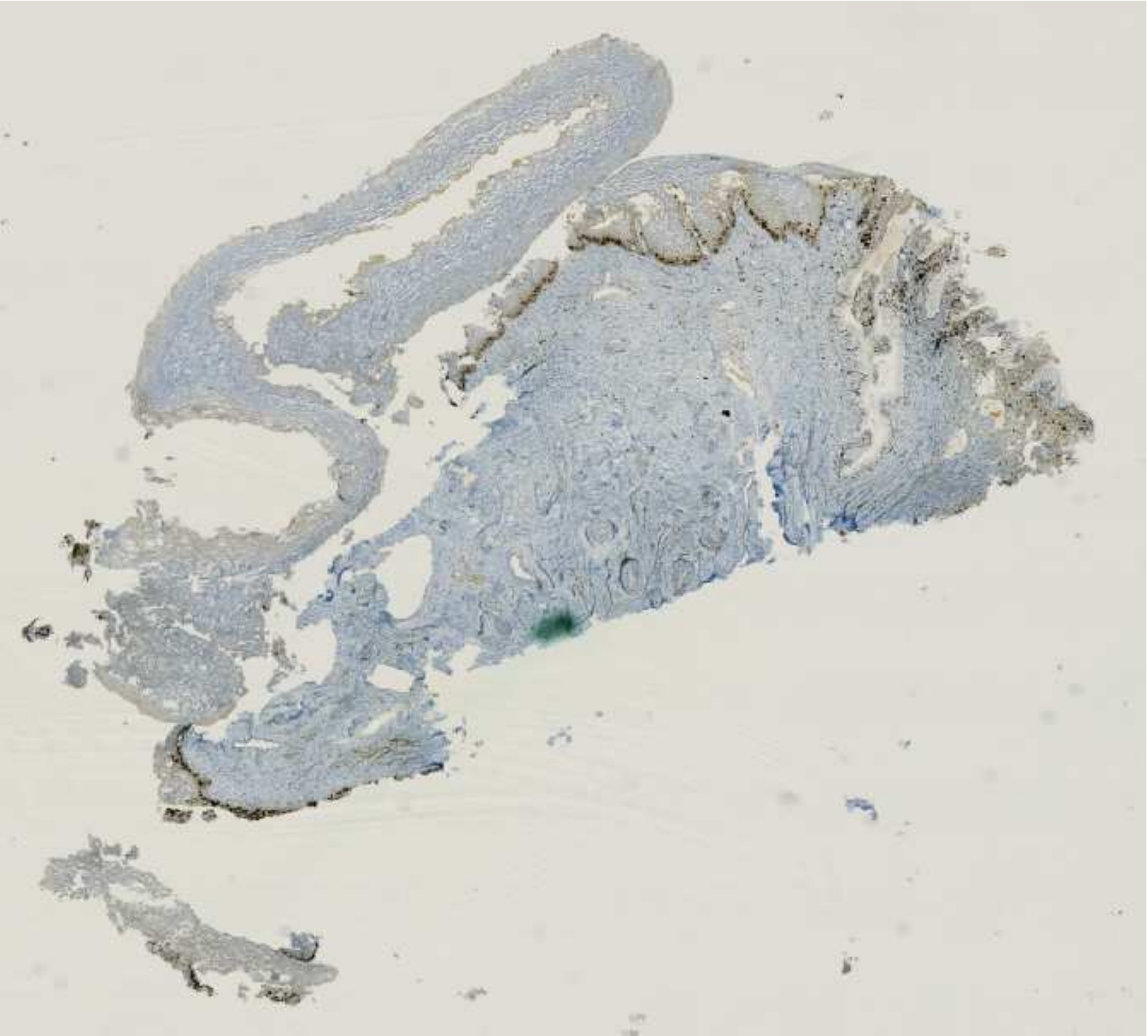}}
  \subfigure[]{
    \includegraphics[width=2.75cm,height=3cm]{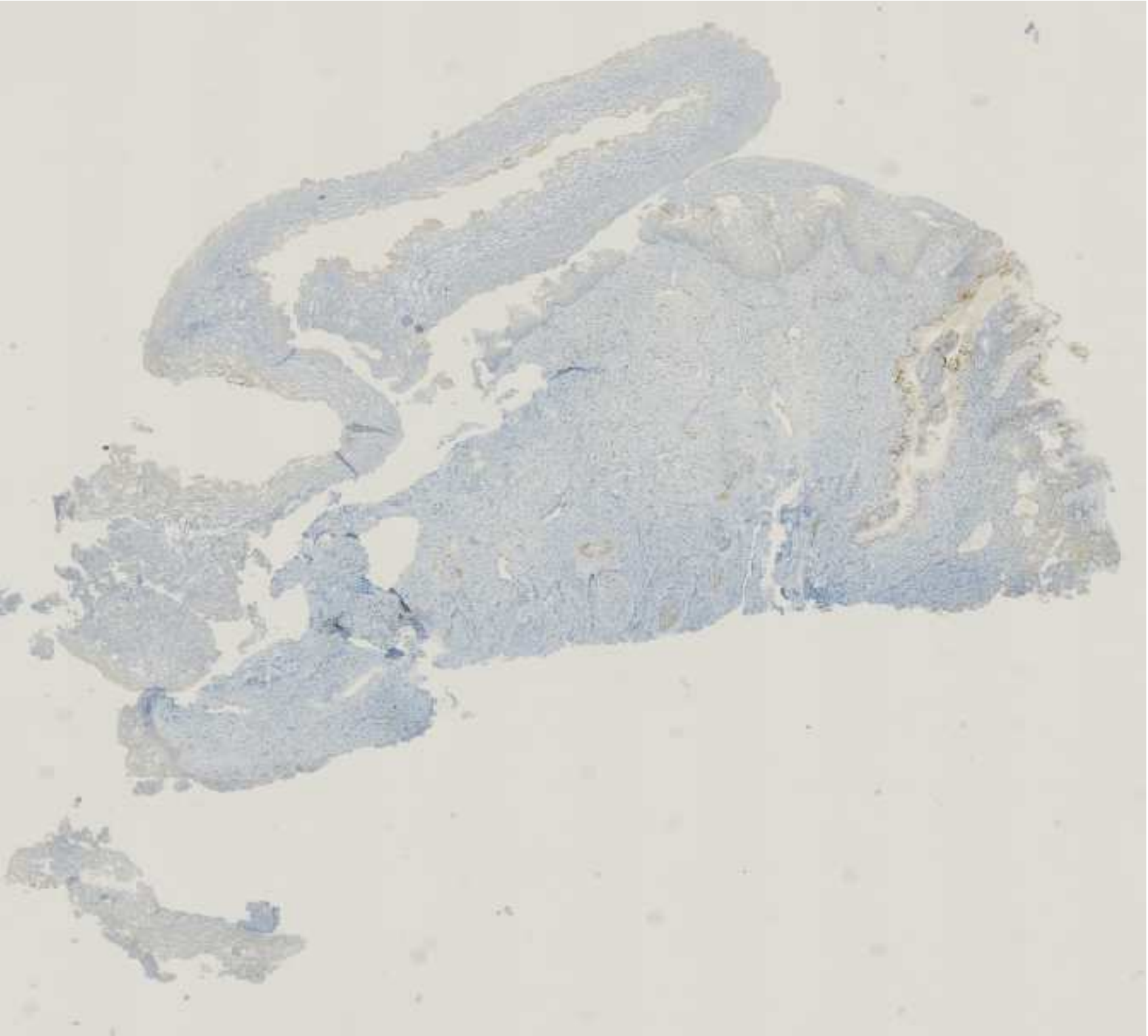}}
  \caption{Image samples obtained from the data set. The sub-figures of (a) and (d) represent the whole slide images which are stained with H$\epsilon$E; the sub-figures of  (b) and (e) represent the same images stained with Ki67 immunohistochemical dye and the sub-figures (c) and (f) shows the same images obtained with p16 immunohistochemical dye.}
  \label{datasetExample}
\end{figure*}

Within the scope of the study, 127 high resolution slides from 54 patients are scanned at the pathology laboratory of IMU Hospital. Fig. \ref{datasetExample} represents the whole slide images obtained from the data set. The images stained with H$\epsilon$ E, Ki67, and p16 immunhistochemical dyes. All high-resolution images are then divided into 957 small epithelium pieces by the pathologist. Each slide in the data set is diagnosed after splitting into smaller epithelial pieces. Totally, 957 epithelial pieces are obtained from the whole slides. 471 of the 957 images  are diagnosed  as  normal,    240  of  them are diagnosed  as  CIN1,  107  of  them are diagnosed  as  CIN2  LSIL,  57  of them are diagnosed as CIN3. 

\begin{figure}[!ht]
\begin{center}
\begin{tabular}{c}
\includegraphics[scale=0.4]{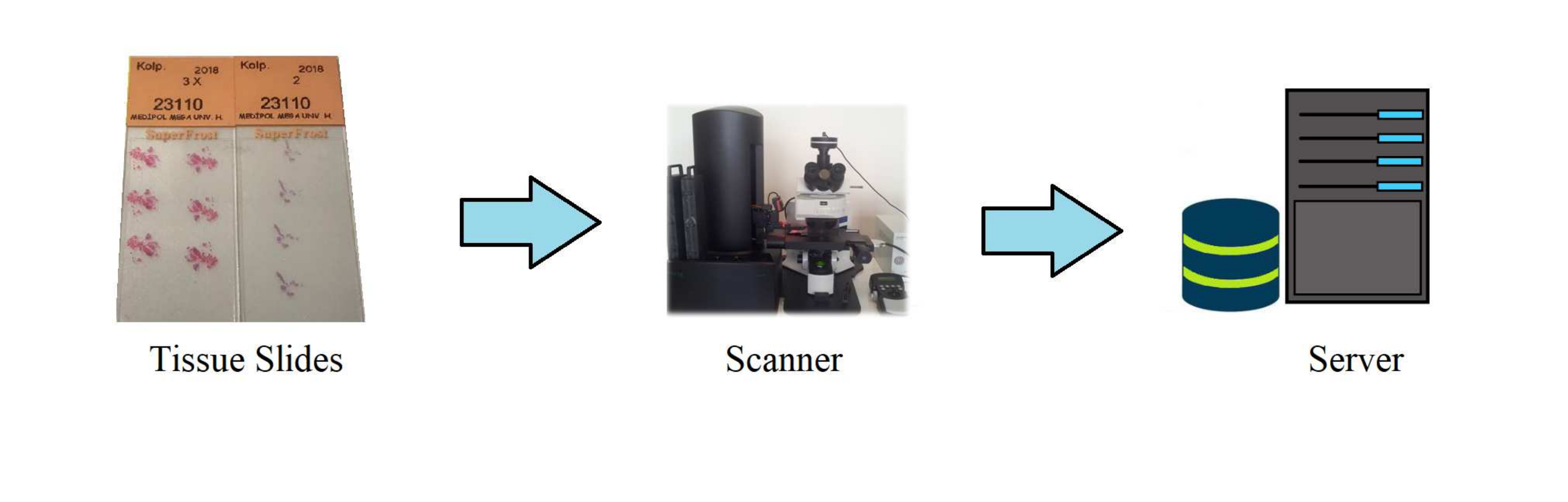}
\end{tabular}
\end{center}
\caption[ ]{Image acquisition system: Tissue slides are scanned using a high resolution scanner. The scanned slides are then transferred to a server to store the images in a file system.}
\label{fig:sakura}
\end{figure}

The images of the hematoxylin and eosin (H$\epsilon$E), p16 and Ki67 preparations are acquired by an off-the-shelf whole slide scanner (See Fig.\ref{fig:sakura}). The scanner has a capability of up to $20\times$ optical and $40\times$ digital zoom. The whole slide images obtained with the high-resolution scanner are transferred to the digital platform to be processed by several image processing techniques and also to be interpreted by the expert pathologists. The images are saved in TIFF format without any loss. The size of the images obtained by the scanner are varied from $10,000 \times 10,000$ to $80,000 \times 80,000$ and there are more than one diagnosis in a single lesion.

\subsection{Ethics Statements} \label{ethics}
Authors confirm that all samples taken from patients were prepared in accordance with the legislation prepared by the Ministry of Health of Turkey and in accordance with international agreements \& European Union standards. All experimental protocols were approved by the Istanbul Medipol University's licensing committee. Informed consent was obtained from all subjects whose tissue samples were used in experiments. In tissue sample collection for data set, there were no subjects under 18. 

\subsection{Annotation and Image Labeling} \label{annotation}
\begin{figure}[!ht]
  \centering
  \subfigure[Sample image]{\includegraphics[width=150px,height=120px]{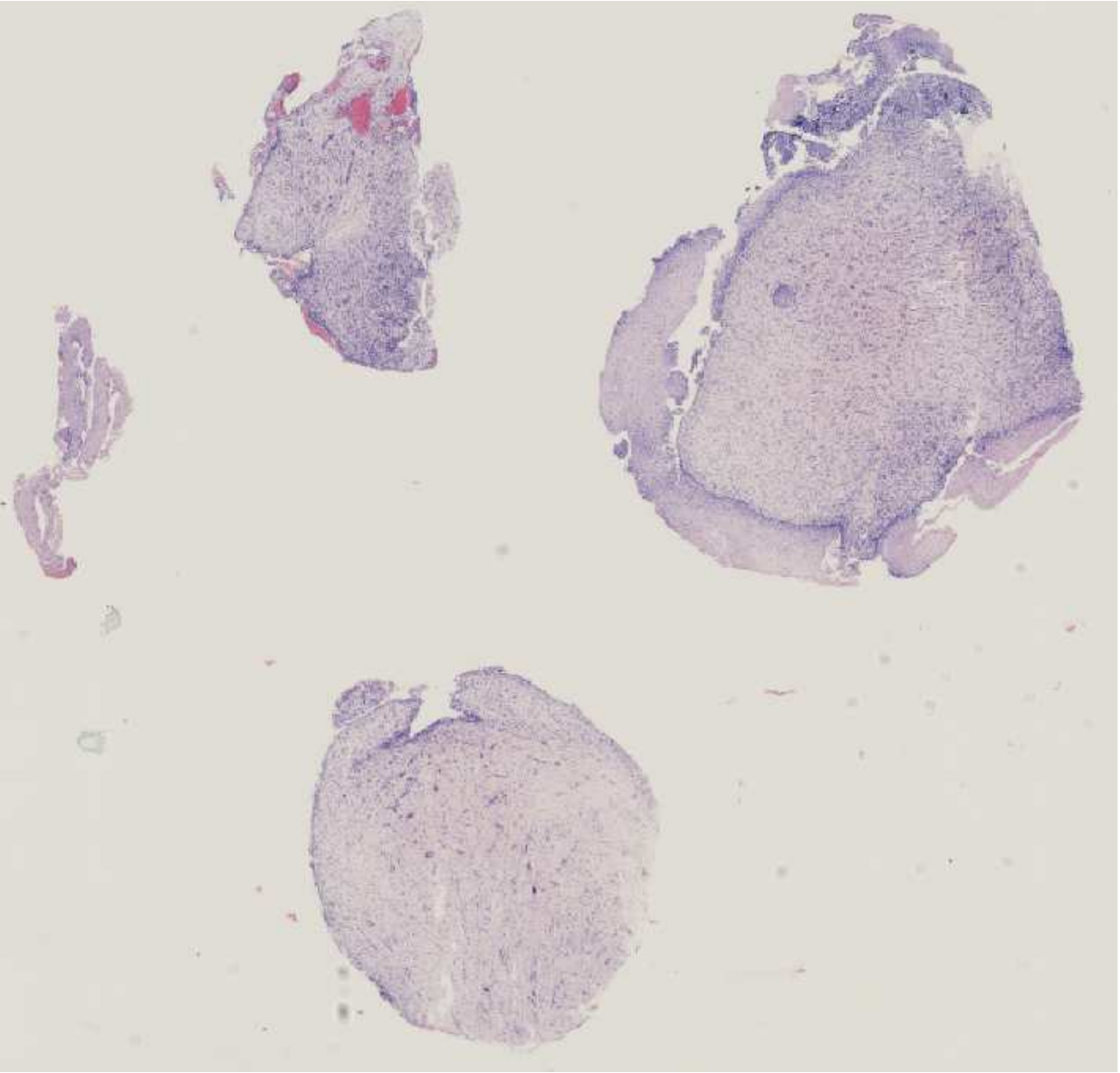}}\quad
  \subfigure[Hot spot (or Region of interest)]{\includegraphics[width=150px,height=120px]{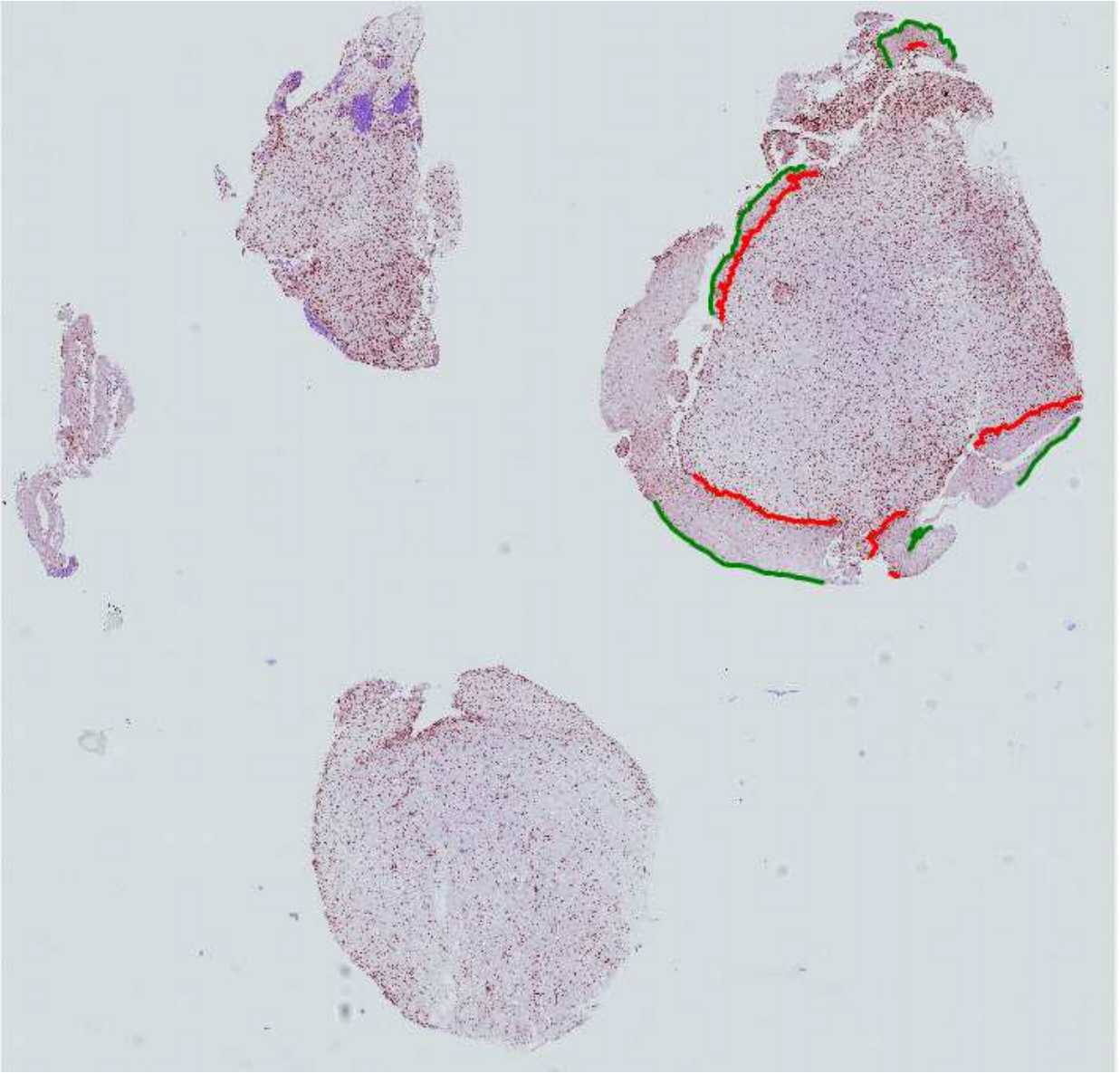}}
  \caption{Annotation and hot spot region extraction. a) Input image obtained from the data set, b) extracted hot spot cervix region for further analysis. The red and green lines drawn around the lesion represents the BM and the upper membrane (UM), respectively.}
  \label{annotationImg}
\end{figure}

A graphical user program is developed within the scope of this study for pathologists to mark/label the basal membranes (BM) and papillae of the cervical epithelium. After marking  the  membranes and the papillae,  the  program  extract  the  hot  spot  region  from  the  background.  Fig. \ref{annotationImg} represents an original input image taken from the data set and a clean image after marking the coordinates of the epithelium. Further image processing and analysis algorithms use this clean image. 

At first, two pathologists made the diagnosis independently for each small epithelial piece (SEP) image patch. A final diagnosis is then made by observing the same lesions stained with p16 and Ki67 immunohistochemical dyes in case of disagreement. According to the final diagnosis, 471 of SEP (\%49.2) are labeled as normal, 240 of them (\%25.1) are CIN1, 107 of them (\%11.2) are CIN2 and 139 of them (\%14.5) are CIN3. However, 150 of large epithelial piece (LEP) (\%46.9) are labeled as normal, 79 of LEP (\%24.7) are CIN1, 34 of LEP (\%10.6) are CIN2 and 57 of LEP (\%17.8) are CIN3 (see Table~\ref{tab:numberofPatchesTripleSystem}). Diagnostic distributions of the SIL-based grading are shown in Table \ref{tab:numberofPatchesBilateralSystem}. Similarly, distribution of final diagnosis in SIL-based grading are as follows: 471 of SEP (\%49.2) are normal, 240 of SEP (\%25.1) are LSIL and 246 of SEP (\%25.7) are HSIL. Similarly, 150 of LEP (\%46.9) are normal, 79 of LEP (\%24.7) are LSIL, 91 of LEP (\%28.4) are HSIL.

\begin{table}[ht]
\centering
\caption{Number of epithelium pieces in each class depending on the CIN-based grading}
\label{tab:numberofPatchesTripleSystem}
\begin{tabular}{|c|l|l|l|l|l||l|}
\hline
\multicolumn{1}{|l }{}                                                   &     & Normal & CIN1 & CIN2 & CIN3 & Total \\ \hline
\multirow{2}{*}{\begin{tabular}[c]{@{}c@{}}CIN-Based\\ Grading\end{tabular}} & SEP & 471    & 240  & 107  & 139  & 957   \\ \cline{2-7} 
                                                                         & LEP & 150    & 79   & 34   & 57   & 320   \\ \hline
\end{tabular}
\end{table}

\begin{table}[ht]
\centering
\caption{Number of epithelium pieces in each class depending on the SIL-based grading}
\label{tab:numberofPatchesBilateralSystem}
\begin{tabular}{|c|l|l|l|l||l|}
\hline
\multicolumn{1}{|l }{}                                                 &     & Normal & LSIL & HSIL & Total \\ \hline
\multirow{2}{*}{\begin{tabular}[c]{@{}c@{}}SIL-based\\ Grading\end{tabular}} & SEP & 471    & 240  & 246  & 957   \\ \cline{2-6} 
                                                                       & LEP & 150    & 79   & 91   & 320   \\ \hline
\end{tabular}
\end{table}

\subsubsection{Inter-Observer Variability} \label{Inter-observerVariability}

Interpretations of morphologic changes representing dysplasia may differ between pyhsicians  or for the same physician in different time intervals. This variety can be interpreted as Inter/Intra-observer agreement / disagreement. Artifacts associated with the biopsy procedure and tangential sections in the microscopic examination are also effective on this variety. Inter- and intra-observer agreement rates are in the range of 0.20 and 0.47 in the literature \cite{mccluggage1996interobserver, mccluggage1998inter}. Regarding to CIN-based grading, SIL-based grading provides higher inter-observer and intra-observer agreement rates. The highest diagnosis diversity is reported between the groups of CIN2, while the lowest is CIN3. The disagreement rates are smaller between normal and CIN1 groups. McCluggage et ~al.~ reported weak inter-observer agreement in the CIN-based grading with Kappa value of 0.2. Although the compatibility rates are reported low, Kappa value is found as 0.3 in the SIL-based grading. Failure to achieve the expected high agreement rates is interpreted by the pathologists involved in the study not being familiar with SIL-based grading. The same test is repeated with the observers who have been experienced to use the SIL-based grading for six months more and new Kappa values are calculated as 0.33 (intra-observer) and 0.47 (inter-observer).  Galgano et al. tried to maximize the agreement rates between the observers with P16 and Ki67 immunohistochemical methods (54). The Kappa value is found to be 0.68 by immunohistochemistry examination while standard hematoxylin and eosin (H$\epsilon$E) detection has the kappa value of 0.47. In the study, it is stated that the low agreement rates associated with diagnostic differences can be increased by using SIL-based grading rather CIN-based grading, or utilizing some immunohistochemical methods aiding diagnosis.

\subsubsection{Final Diagnosis} \label{finaldiagnosis}

Immunohistochemical examinations are used as an assistive method to obtain the diagnosis in case the morphological features are not clearly interpreted. P16, Ki67 and ProExC are the most widely used immunohistochemical studies for cervical precursor lesions \cite{melo2016expression, galgano2010using, guo2011efficacy, lim2016efficacy, ozaki2011biomarker}. Staining pattern with p16 is important in immunohistochemical evaluation, and block-like and strong staining demonstrates HrHPV association, with at least 1/3 of the epithelium (7,56,58). Ki67 is an indicator of proliferation. Positivity may also be seen in other proliferating cells such as inflammatory cells as in keratinocytes. For this reason, it must be interpreted carefully in the presence of inflammation. ProExC is similar to Ki67 in terms of being a proliferation indication and its staining type. P16 and Ki67 are frequently used in routine practice. HrHPV-associated lesions show strong “nuclear” or “nuclear and cytoplasmic”, block-like staining with P16. The squamous metaplasia, atrophy, reactive regenerative changes that appear in the SIL discriminator pattern show a negative staining pattern. While Ki67 normally stains parabasal cells, positivity is also observed in higher epithelial sections in relation to the grade of dysplasia in SIL. Reactive cells and inflammatory cells may also exhibit immunoreactivity, it should be very careful when interpreting these tissues.

\subsection{Morphometric Feature Extraction and Tissue Classification} \label{ProposedMethod}

In this study, a morphological analysis based feature extraction method is used for the grading of cervical cancer precursor lesions. The processing steps followed in the study is represented in Fig. \ref{fig:processingSteps}. The first row of the diagram describes the pathological pre-process which is handled in the pathology laboratory. Whole slide scanning and the filing process mentioned  in the second row is done by the medical researcher and the computer scientist in collaboration. This section describes the CADAS framework which is mentioned in the third row.

\subsubsection{Creating the Small Epithelial Pieces (SEP)}
\begin{figure}[ht]
\begin{center}
\begin{tabular}{c}
\includegraphics[scale=0.5]{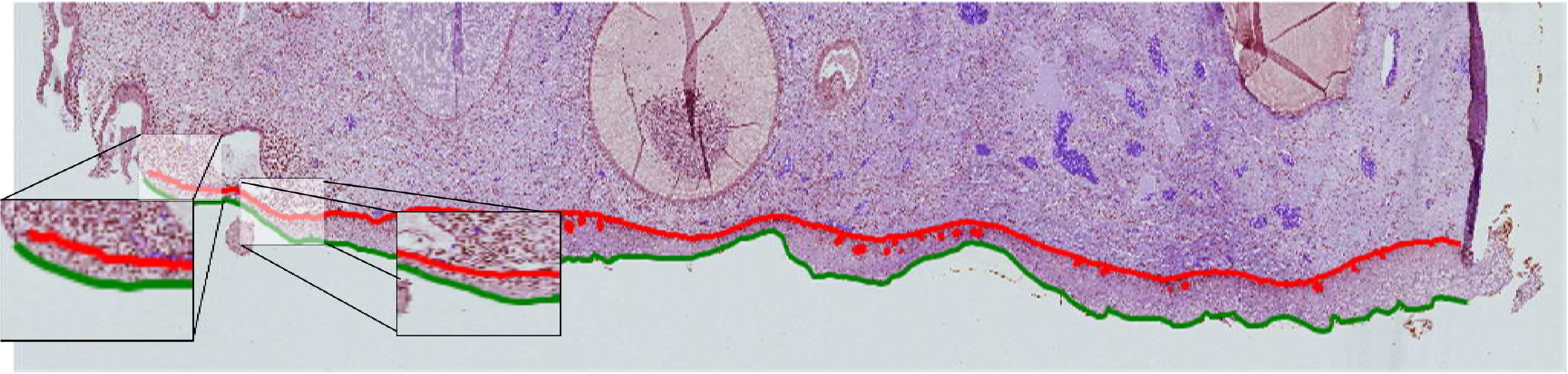}
\end{tabular}
\end{center}
\caption[]{A high resolution histopathological image example obtained from the data set and the SEP cropped from that image.}
\label{fig:SamplePatchExtraction}
\end{figure}

In Fig.~\ref{fig:SamplePatchExtraction}, the red line corresponds to the coordinate information of the BM, while the green line corresponds to the coordinate information in the upper membrane. Determining the basal and upper membrane coordinates allows to know in which region of the epithelium the cells are located. After the region of interest (squamos epthitelium) has been obtained, an interface developed within the scope of the study is used to divide the whole epithelium to SEP which can be assumed equal in length (see Fig.~\ref{fig:r_g_papillae}).

The image patches which are analyzed in this study are represented in Fig.~\ref{fig:r_g_papillae}. The coordinates of basal and upper membranes of the epithelium are marked by the pathology experts with the use of a graphical interface. Coordinates data information of the papillae which represented with yellow line are also stored in separated files.

\subsubsection{Obtaining Cells by Simple Linear Iterative Clustering(SLIC) Superpixels Segmentation Algorithm}
After small epithelial pieces are obtained, the high resolution histopathological images are ready for further analysis. First,  a median filter of dimension $9\times9$ is applied to the image to remove the artifacts without effecting the boundaries. Then, cellular structures have been obtained by simple linear iterative clustering (SLIC)  superpixels segmentation algorithm, which is one of the methods that have not been widely used yet in histopathological images in recent years. This method performs the segmentation process based on the color similarities and neighbour relations of the pixels in the image \cite{achanta2012slic}. The grid size $N$ is expressed as

\begin{equation}
     N = \sqrt{\frac{w \times h}{k}}
    \label{equation4}
\end{equation}
where $k$ is the number of superpixels for a given input image, $w$ and $h$ represent the width and height of the given image patch, respectively. The euclidean distance of the related pixel to the superpixel center is 

\begin{equation}
    d_{rgb} = \sqrt{(r_{j}-r_{i})^2 + (g_{j}-g_{i})^2 + (b_{j}-b_{i})^2}
    \label{equation1}
\end{equation}

\noindent
where, $i$ represents the value to be clustered and $j$ represents the center pixel. Here, $ r, g$ and $b$ represent the brightness values of red, green, and blue color of the respective pixels. RGB color space is used in this study instead of using Lab color space as mentioned in Achanta et al\cite{achanta2012slic}. The Eq.~\ref{equation2} also represents the distance of the coordinates of each pixel to the related cluster center:

\begin{figure}[ht]
\begin{center}
\begin{tabular}{c}
\includegraphics[scale=0.45]{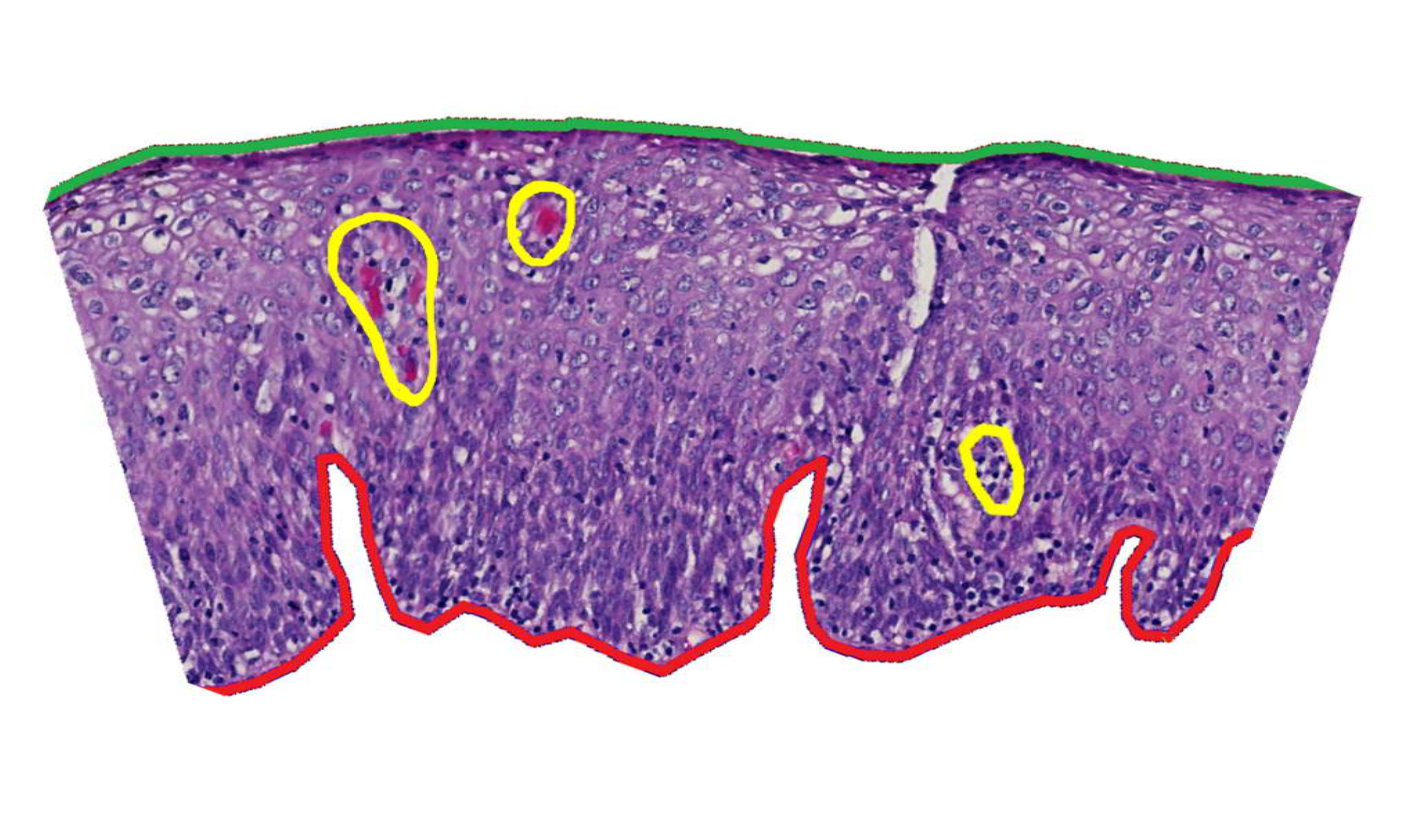}
\end{tabular}
\end{center}
\caption[]{Sample SEP image patch which includes basal, upper membrane and papillae. The grading is done taking this structures into consideration by the pathologists. The red line, green line and yellow circle define the BM, upper membrane and papillae coordinates information, respectively.}
\label{fig:r_g_papillae}
\end{figure}

\begin{equation}
    d_{xy} = \sqrt{(x_{j}-x_{i})^2 + (y_{j}-y_{i})^2}
    \label{equation2}
\end{equation}

\noindent
where, $ x_{j}$ and $y_{j}$ are the horizontal and vertical coordinate information of each center pixel, and $ x_{i}$ and $y_{i} $ values are the coordinate information of each pixel to be clustered:

\begin{equation}
    d_{s} =  d_{rgb} + m/(N \times d_{xy})
    \label{equation3}
\end{equation}

\noindent 
the value of $ d_{s} $ is the sum of the (x, y) plane distance normalized by the grid interval N and the RGB distance. Here, normalization is done so that the calculation of the coordinate information does not directly affect the brightness interval. The value of $m$ is defined to set the compactness of superpixels.

\begin{figure}[!htbp]
  \centering
  \subfigure[SEP image patch]{\includegraphics[scale=0.65]{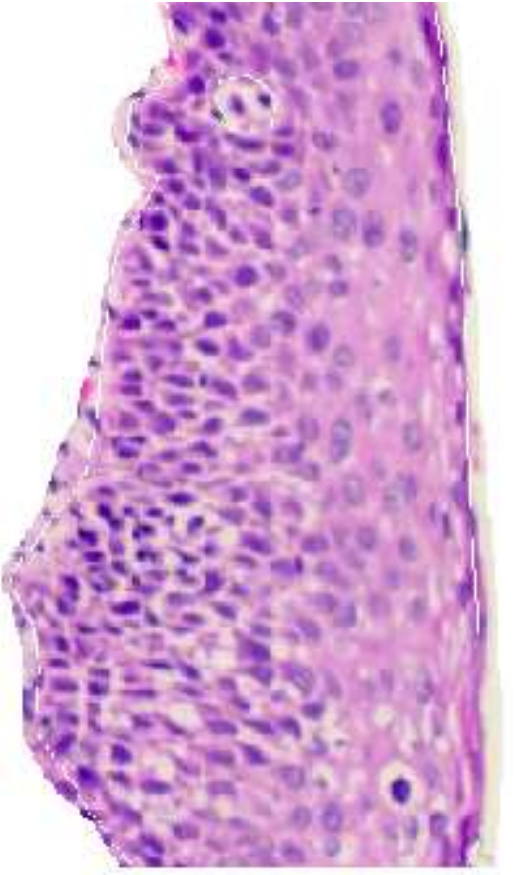}}\quad
  \subfigure[Overlay of superpixels]{\includegraphics[scale=0.65]{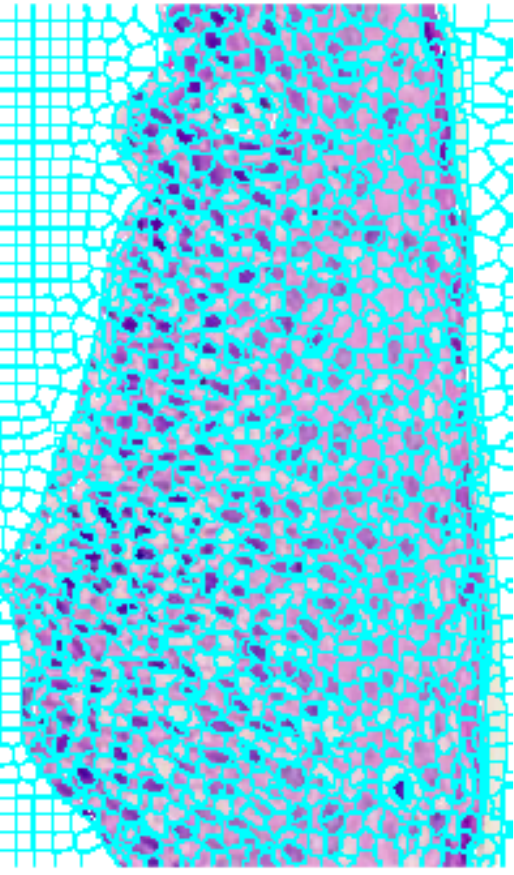}} \quad
  \subfigure[The pre-segmented SEP image patch ]{\includegraphics[scale=0.65]{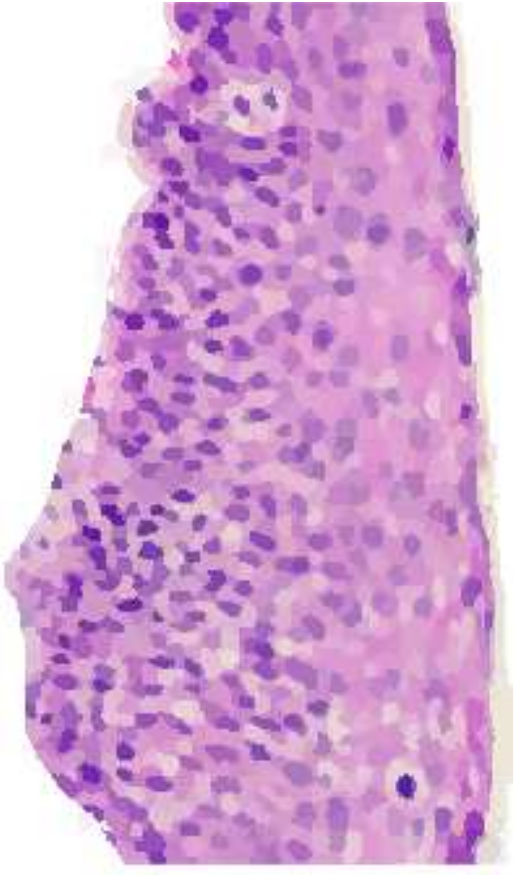}}
  \caption{Implementation of SLIC superpixels segmentation algorithm to a sample image patch  a) SEP image patch, b) Overlay of 3000 superpixels on the related image patch, (c) Resulting pre-segmented image obtained after applying SLIC method.}
  \label{suerpixelSegmentation}
\end{figure}

\begin{figure*}[ht]
  \centering
  \subfigure[]{
    \includegraphics[scale=0.45]{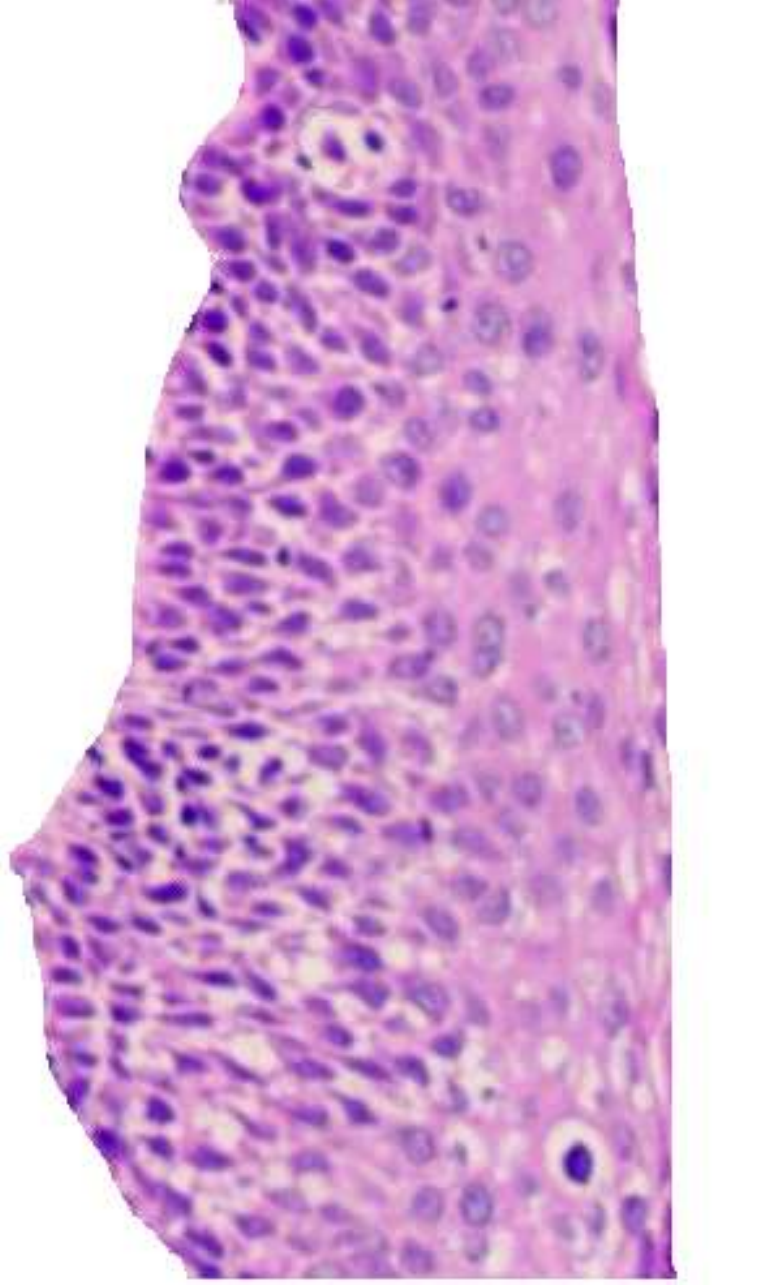}} 
  \subfigure[]{
    \includegraphics[scale=0.45]{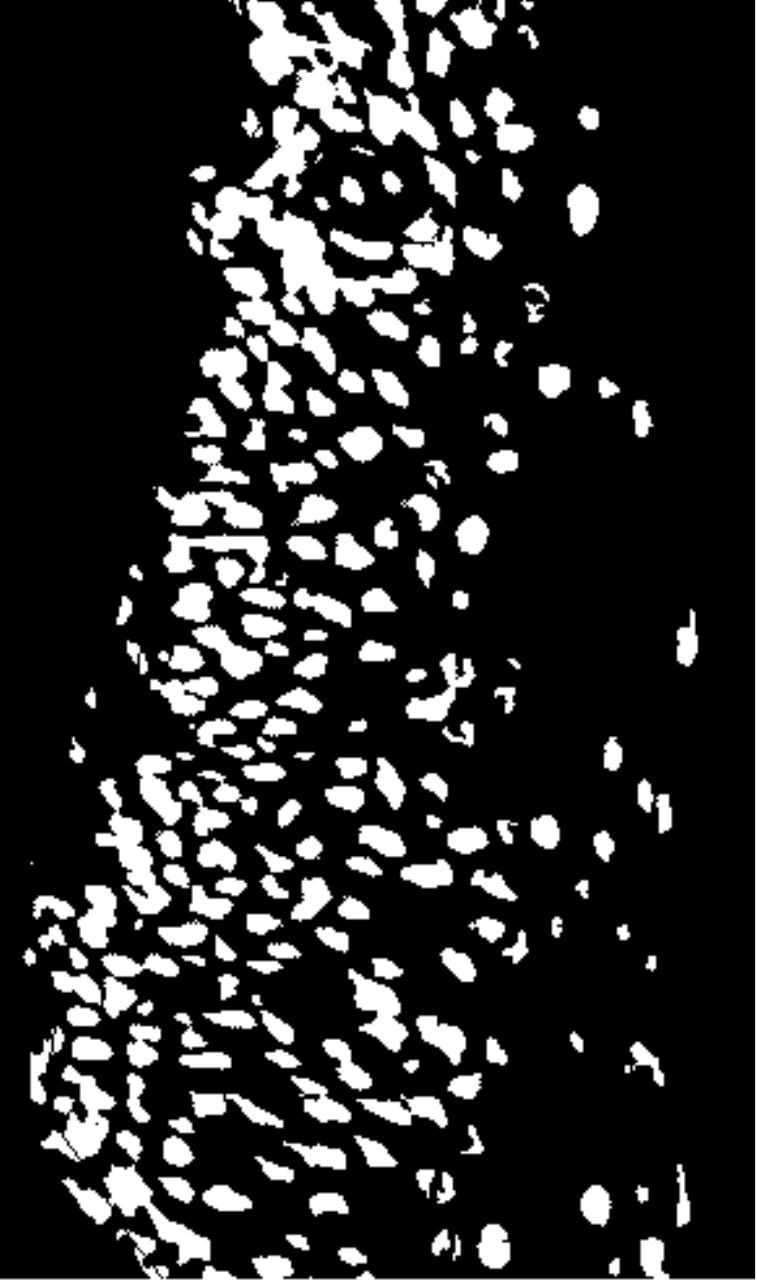}}
  \subfigure[]{
    \includegraphics[scale=0.45]{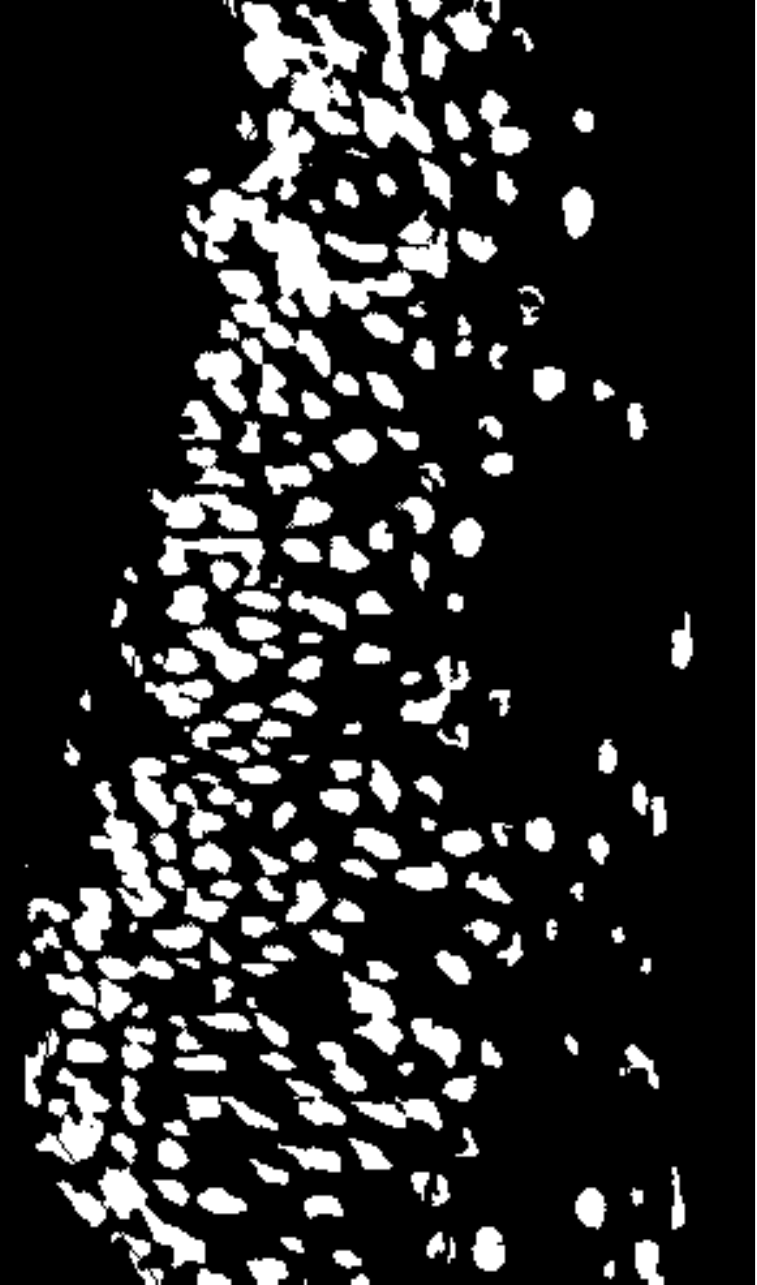}}

  \caption{Final segmentation result of a sample SEP image  patch (a). given input image (b) segmentation of the image (c) final binary image after post processing.}
  \label{fig:finalresultafterpostprocessing}
\end{figure*}

\begin{algorithm}[ht]
\caption{Center Estimation and Distribution Generation}\label{alg:C3}
\hspace*{\algorithmicindent}\textbf{Input} $\mathbf{I}_{dist}\in\Re^{N\times M}$\Comment{Distance transform of binary image}\\
\hspace*{\algorithmicindent}\textbf{Output} $\mathbf{C}\in\Re^{K\times 2}$ \Comment{Cell center matrix} 
\begin{algorithmic}[1]
\vspace{3pt}
\Procedure{Local Maxima Finding}{} 
\For{$\forall(i,j)\in \mathbf{I}_{dist}$} 
\vspace{5pt}

\State $\mathbb{S} = \left\{(u,v)\hspace{3pt}|\hspace{3pt}(i-u)^2+(j-v)^2 \leq r^2 \right\}$
\vspace{5pt}

\If{ $\arg\max\left\{\mathbf{I}_{dist}(\mathbb{S})\right\} == (i,j)$} 
\vspace{5pt}

\State $\mathbf{C}_{k,:} \gets [i,j]$
\vspace{5pt}

\EndIf

\EndFor

\label{euclidendwhile}
\State \textbf{return} $\mathbf{C}$

\EndProcedure
\end{algorithmic}

\vspace{8pt}
\hspace*{\algorithmicindent} \textbf{Input} $\mathbf{I}_{dist}\in\Re^{N\times M}$\\
\hspace*{\algorithmicindent} \textbf{Output}  $\mathbb{H}$ \Comment{Multiplexed Coordinates Set} 
\begin{algorithmic}[1]
\vspace{3pt}
\Procedure{Test C3}{} 
\State $\mathbb{H} \gets \emptyset$ 
\For{$\forall(i,j)\in \mathbf{I}_{dist}$} 
\vspace{5pt}
\State $ \mathbb{H} \gets \mathbb{H} \hspace{5pt}\cup \hspace{5pt} \alpha \otimes [i,j] $  
\Comment{$\otimes$ : multiplexing operator} 

\vspace{7pt}
\EndFor
\State \textbf{return} $\mathbb{H}$
\EndProcedure

\end{algorithmic}
\end{algorithm}

According to the SLIC method applied in this study, the cellular structures become more compact and can be separated from the background when each obtained SEP image is expressed with 3000 superpixels. A cellular structure in cervical precursor lesion is approximately $20\times20$. A crucial point to note here is that the superpixels' sizes should not exceed the size of cellular structure. As can be calculated from the Eq.\ref{equation4}, choosing at least 2000 superpixels will guarantee most of the superpixels to not exceed the size of a cellular structure. Less number of superpixels cause overlapping cells. It can be quite difficult to distinguish cellular structures, especially those close to the BM. Since the superpixels which represent the cellular structures are darker than the superpixels which represent the background, the elimination of the unwanted pixel groups (fat-like tissue) above a certain threshold level clears the background. At this stage, small artifacts similar to the cellular structures and some inflammation can remain with the cells as a foreground information. These structures can be eliminated with a morphological size operation that can be applied to the binary image after segmentation stage. Final segmentation result of a sample SEP image patch obtained from the data set is shown in Fig. \ref{fig:finalresultafterpostprocessing}.

\subsubsection{Handling the Overlapping Cells Problem}
Following the morphological operations, overlapping cells are separated. In the literature, the problem of overlapping nuclei in a plurality of nucleus segmentation studies has been encountered. Because the presence of overlapping cell structures significantly reduces the success of CADAS. Solving the problem of overlapping nucleus at this point is very crucial as a significant contribution in this area. In our study, it is observed that after the segmentation process, there are a large number of overlapping nucleus structures, especially around the BM. 

\begin{figure*}[!ht]
    \centering

    \subfigure[Binary Cell Image]  
        {\includegraphics[width=.30\textwidth]{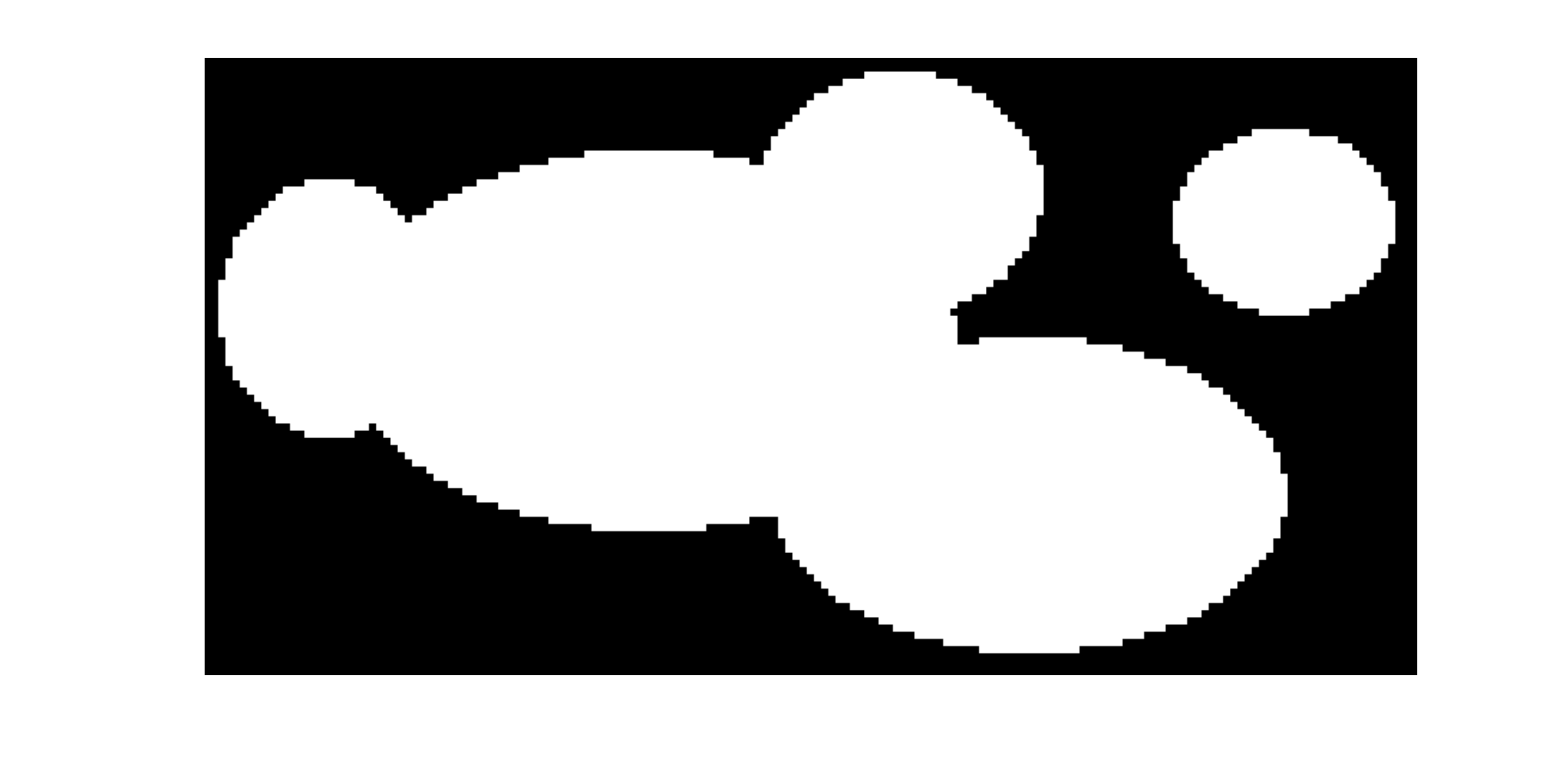} \label{fig:sub1}} \quad
    \subfigure[Distance Transform]
        {\includegraphics[width=.30\textwidth]{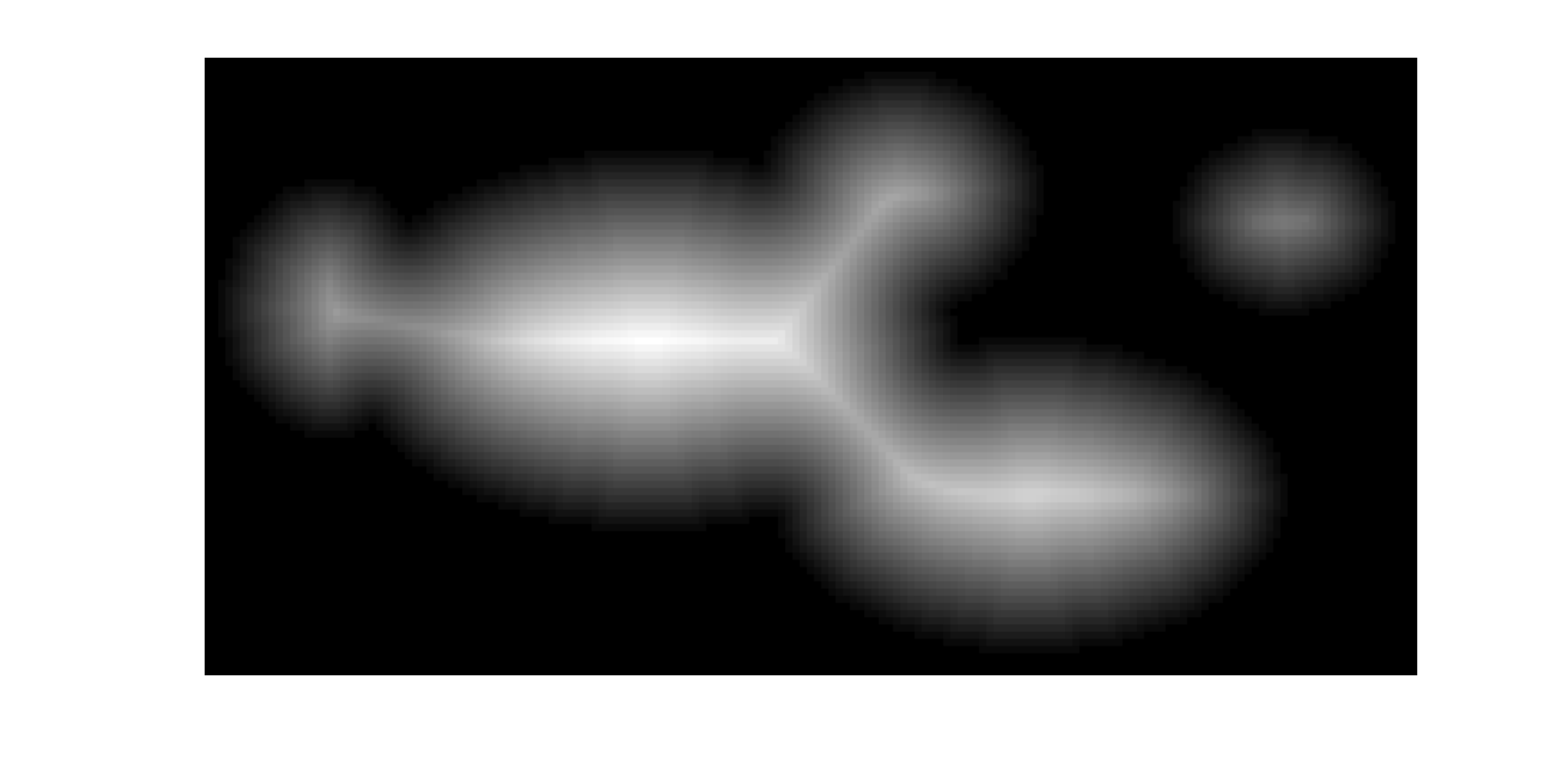} \label{fig:sub2}} \quad
     \subfigure[Center Estimation]
        {\includegraphics[width=.30\textwidth]{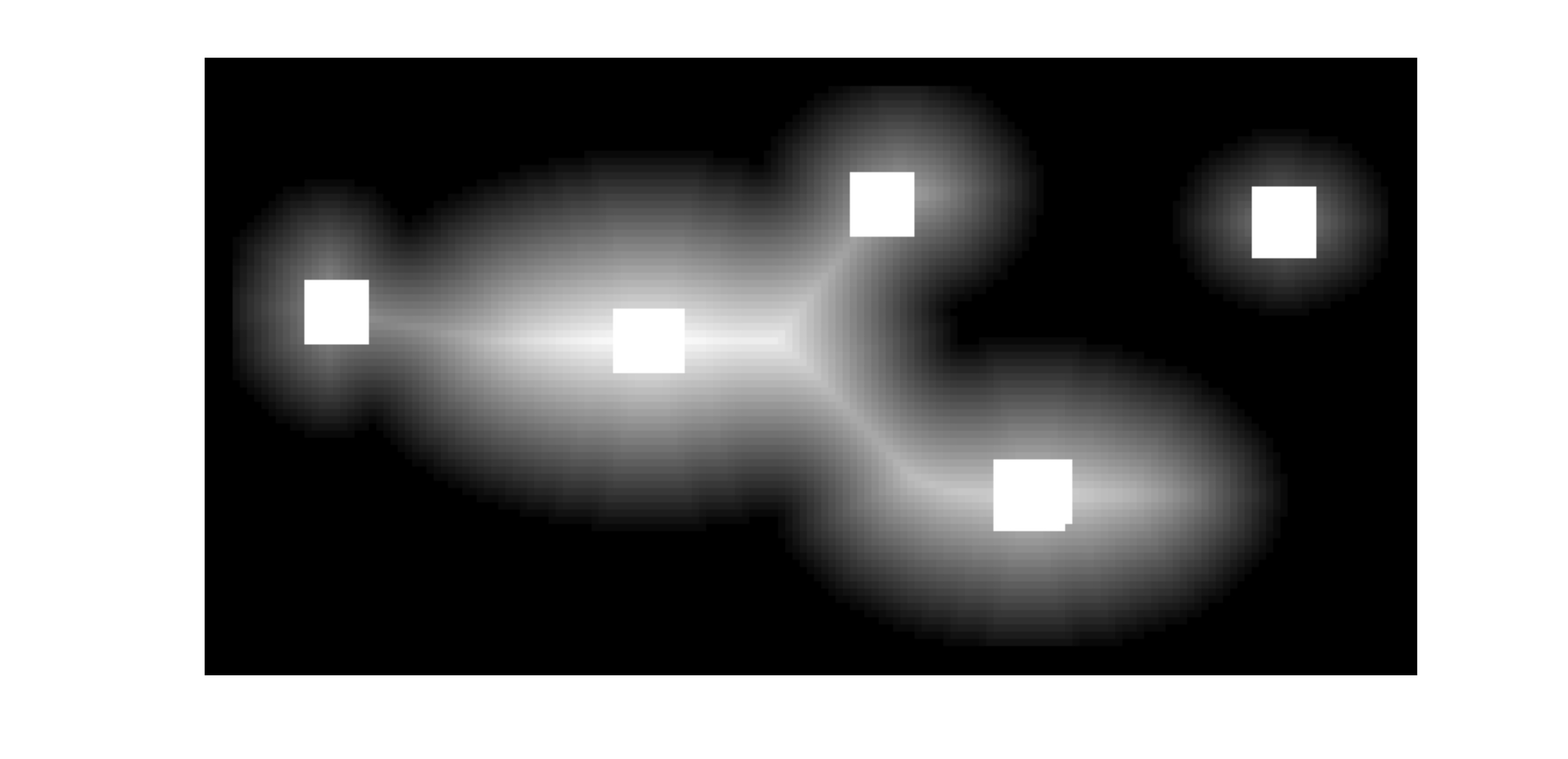} \label{fig:sub3}}

    \caption{Cells taken from the tissues are in often overlapped form. For the solution of this problem, it is important that the cell centers are firstly correctly estimated. (a) Binary mask of overlapping cell heaps (b) Distance transform of binary mask (c) Finding local maximums.}
    \label{fig:SignVar1}
\end{figure*}

Overlapped cells are intensively present on the SEP image patches. The problem of overlapping of cells should be handled in order to obtain the morphological characteristics of the cell nuclei. Binary images segmented by using SLIC algorithm usually consist of small cellular-like noisy parts. These unwanted small pixel groups are eliminated by an automatic method, which clears pixel groups smaller than 50 pixel. Therefore, the circumference of the cells is also quite rough after the segmentation process. As shown in Fig. \ref{fig:sub1}, the closing process has been applied to make the binary image more compact. In order to obtain cell centers, the distance transform is applied to get local maxima shown in Fig~\ref{fig:sub3}. Local maxima are estimated by using Algorithm-1.

The ellipse form is able to model the cell shapes mathematically well. Thereby, Gaussian Mixture Model (GMM) is one of the best candidate for ellipse fitting over cell heaps. GMM is one of the most common algorithms for statistical data modeling \cite{najar2017comparison}. The main purpose of the algorithm is to express the distributions as the sum of the weighted Gaussian mixtures (see Eq. (\ref{eq:gmm})). $\mathcal{N}(\mathbf{x} | \mathbf{\mu}_i , \Sigma_i)$ and $\phi$ intend normal distribution which has mean $\mu_i$ and covariance matrix $\Sigma_i$  and its weight parameter.

\begin{gather}  \label{eq:gmm}
    p(\mathbf{x})      = \sum_{i=1}^{K}\phi_i \mathcal{N}(\mathbf{x} | \mathbf{\mu}_i , \Sigma_i) \\
  \sum_{i=1}^{K}\phi_i = 1    
\end{gather}

\vspace{5pt}

\begin{figure}[ht]
\begin{center}
\begin{tabular}{c}
\includegraphics[]{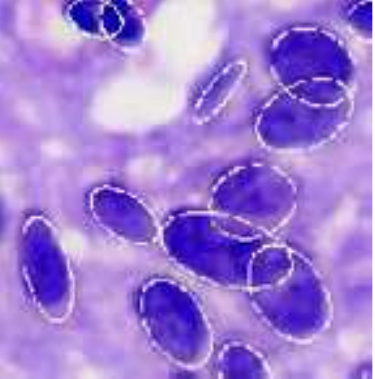}
\end{tabular}
\end{center}
\caption[]{Separating overlapping cell after the segmentation process.}
\label{fig:overlap_nuclei}
\end{figure}

\begin{figure*}[!ht]
    \centering

    \subfigure[Uniform Distribution]  
        {\includegraphics[width=.42\textwidth]{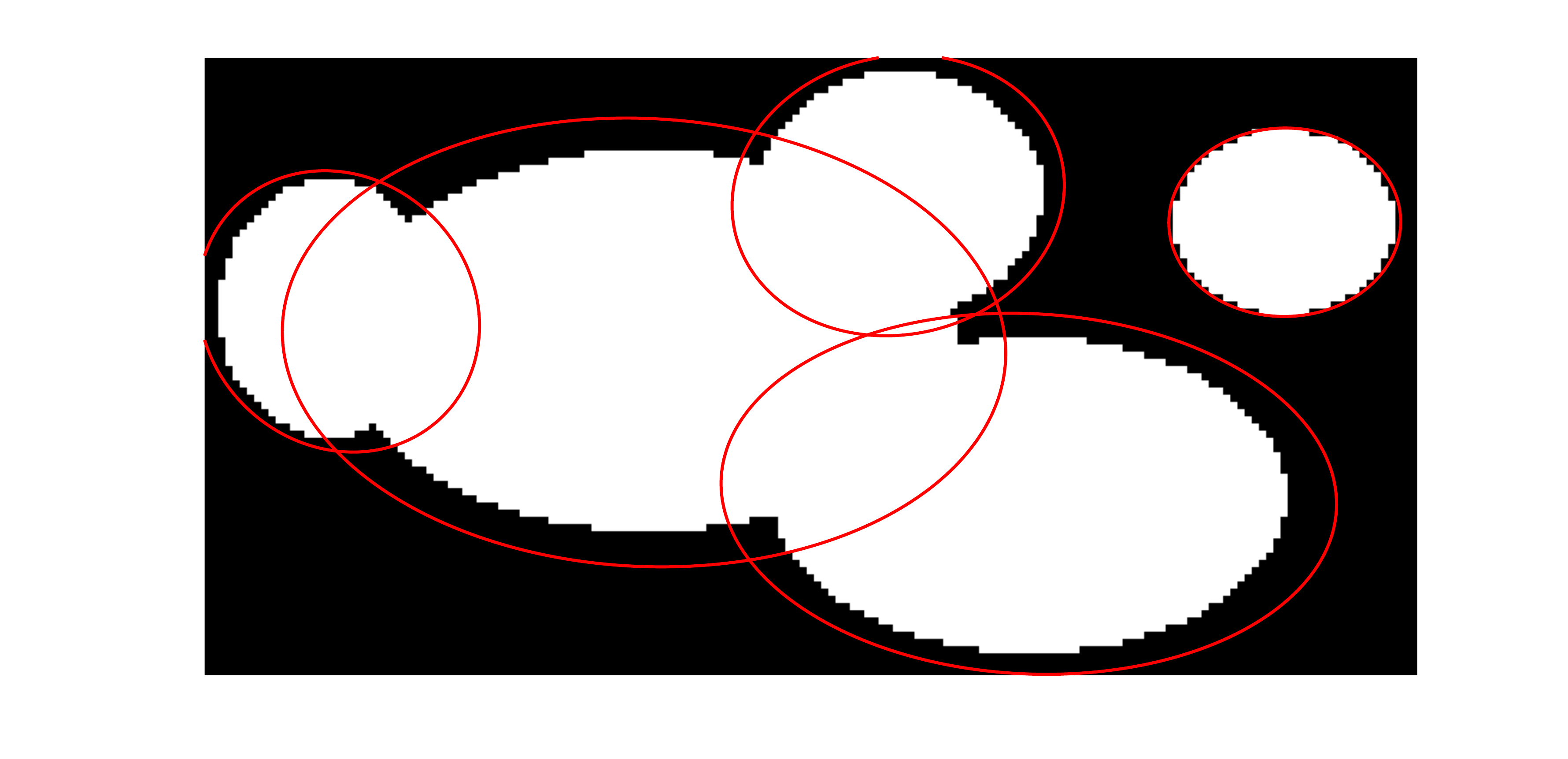}} \quad
    \subfigure[Normal Distribution]
        {\includegraphics[width=.42\textwidth]{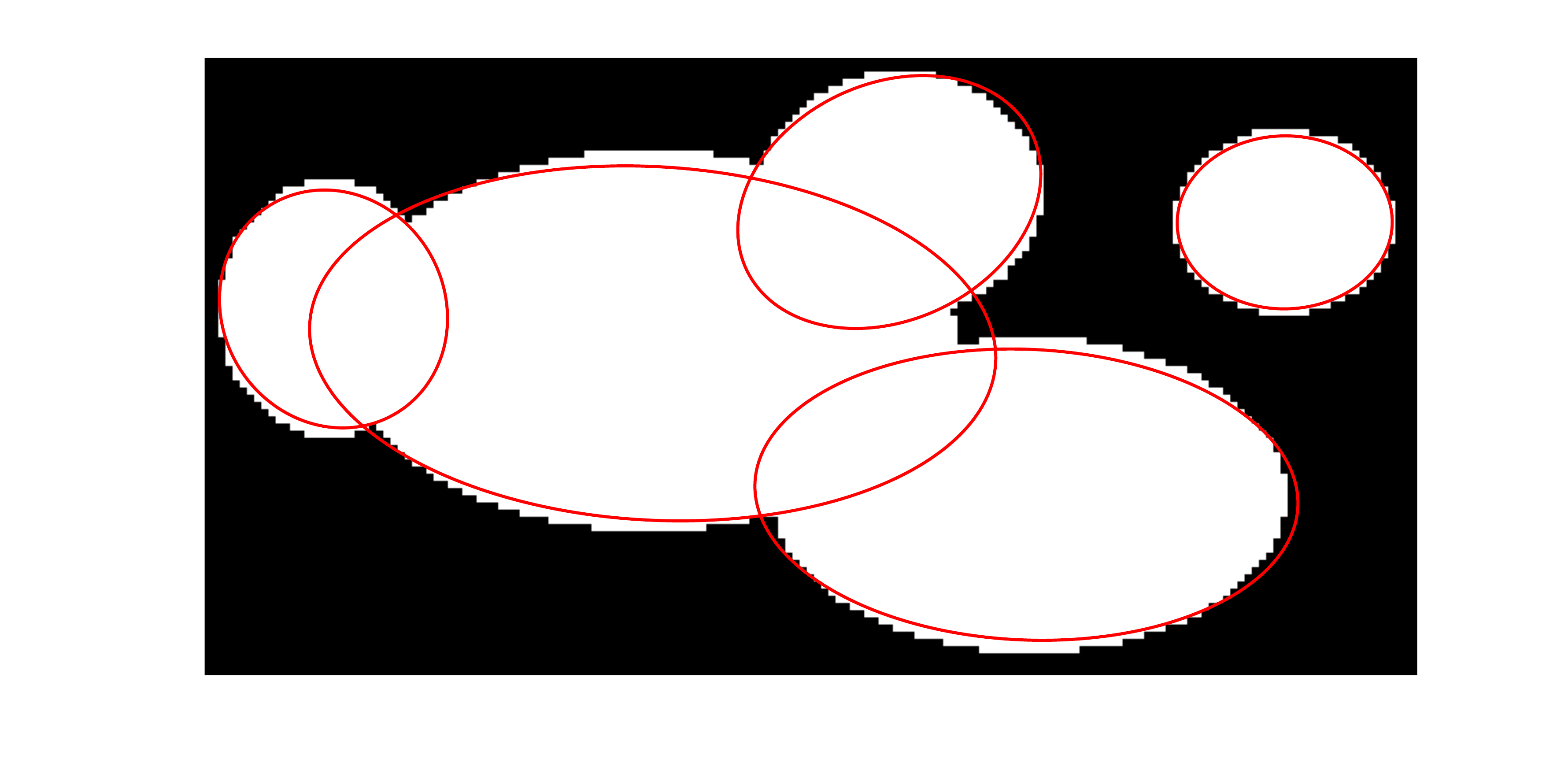}} \quad  

    \caption{Generation of ellipses based on uniform and normal distribution after the estimation of center locations: The cell population modeled as the normal distribution is more suitable for the GMM algorithm than the uniform distribution.}
    \label{fig:SignVar}
\end{figure*}

Fig. \ref{fig:overlap_nuclei} shows an example of overlapping nucleus and how these overlaps are resolved  in a small patch of the image obtained from the data set. The basic structure of the algorithm that determines the cell overlapping is based on the determination of local maxima from the distance of the cell centers to the boundaries. Once the cell centers are determined, the distance transformation yields the value of the multiplexing for each pixel. The processing steps applied for center estimation, distribution generation and multiplexing are given in Algorithm-1. $\mathbf{C}$ and $\mathbb{H}$ refer to row-wise cell centers matrix, and multiplexed coordinate set, respectively.The $\alpha$ parameter obtained from the distance transform indicates how many times the corresponding coordinate will be repeated in the set $\mathbb{H}$. Thereby, the distance value of the pixels away from the border are higher, so the amount of these in $\mathbb{H}$ will be more. Pixel distribution becomes more suitable for GMM. For example, if each pixel is far away from a boundary, the coordinate information of the related pixel is multiplexed. By applying GMM on the obtained multiplexed coordinate distribution, suitable ellipses are obtained for each cell (see Fig. \ref{fig:SignVar}).  

\subsubsection{Obtaining the Morphological Features of Each SEP }

\begin{figure*}[ht]
  \centering
  \subfigure[]{
    \includegraphics[scale=0.5]{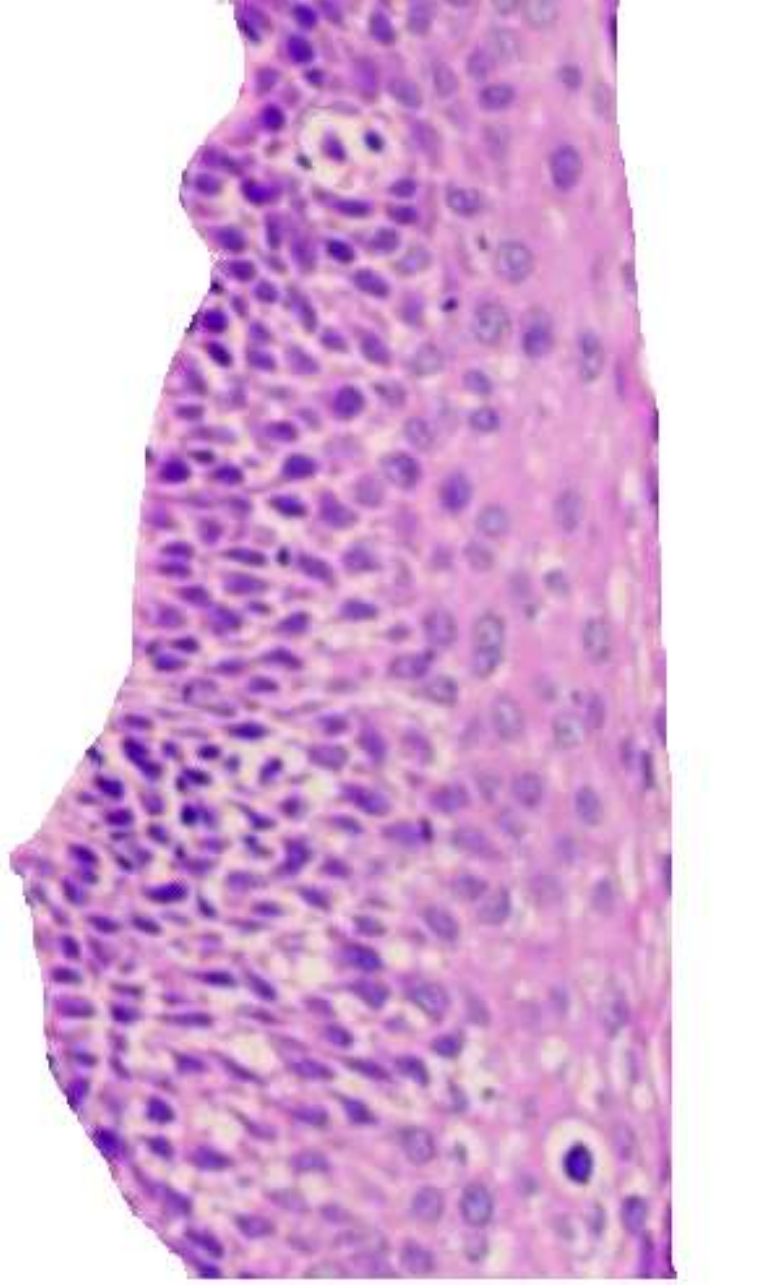}}
  \subfigure[]{
    \includegraphics[scale=0.5]{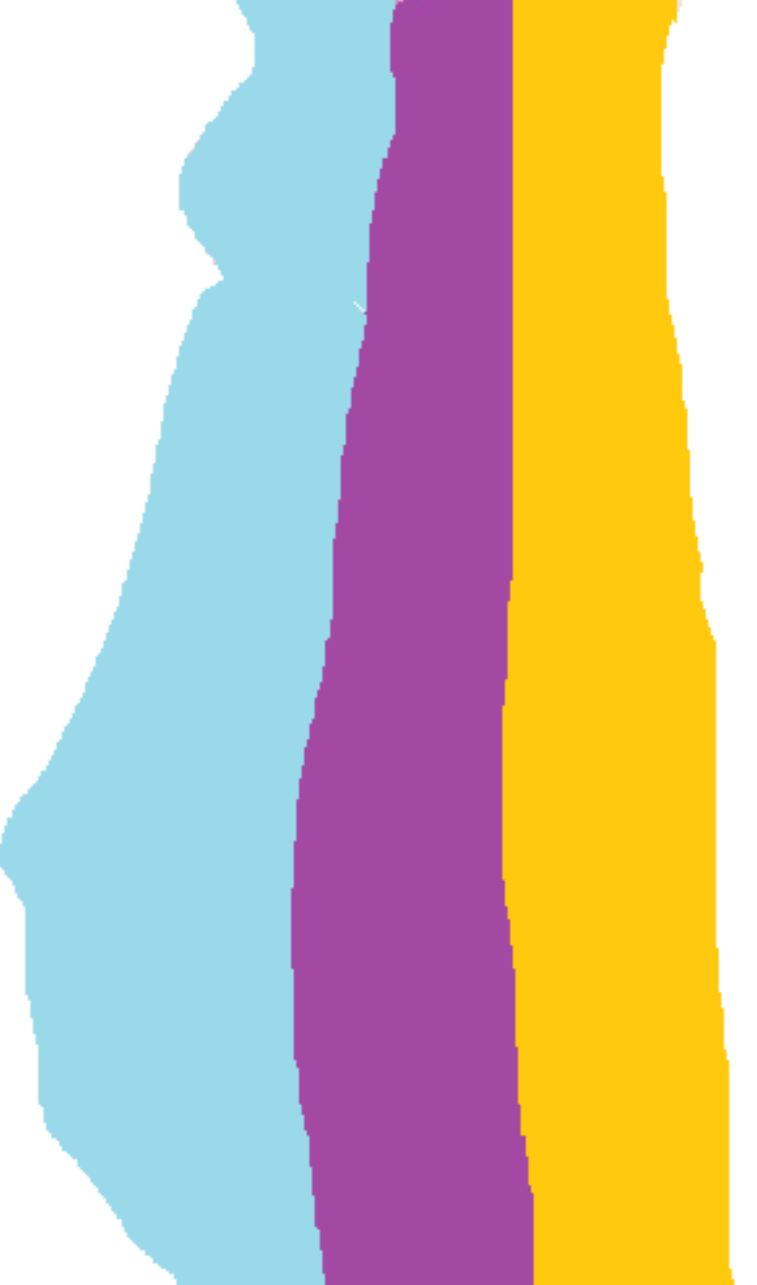}}
  \caption{ (a) Sample image obtained from the data set and (b) layered result of the same image into three section from basal to upper membrane. The distance of each pixel to the basal and upper membrane is calculated by using the coordinate information. The distance from the pixel coordinate to the each membrane indicates its belonging region.}
  \label{fig:regionDivision}
\end{figure*}

After the cell segmentation and elimination of the cell overlap problem, several morphological features of each cell are extracted. Table \ref{tab:morphologicalFeatures} represents the morphological features extracted for grading each SEP image patch in this study. Average nucelus area (ANA), average cytoplasm area (ACA), nucleus - cytoplasm area ratio (NCR), nucleus perimeter (NP), border irregularity (BI), hyperkromasis index (HI), and polarity loss index (PLI) are the features represented from the first row to the end, respectively. ANA defines the average nucleus area while ACA is the average cytoplasm area. NCR describes the division results of nucleus ratio to the cytoplasm ratio. NP is the average length of the border pixels of nucleus. BI is the divison of the surrounding length of each pixel to the ellipse that fit each nuclei. The value which represent the hyperkromasis of cell is calculated by taking standard deviation of the pixel intensity values of the cellular structure. PLI is determined by calculating the magnitude of each cellular structure to the BM. All the features are extracted for each region shown in Fig. \ref{fig:regionDivision}

Since the emphasized morphological features change depending on their distance to the basal and the upper membranes, each image segment is divided into three main regions as represented in Fig. \ref{fig:regionDivision}. Then, morphological features related to each region are extracted and stored for further analysis in grading the SEP image patch.

\begin{table}[!ht]
\centering
\caption{List of morphological features extracted in the proposed tissue classification method}
\label{tab:morphologicalFeatures}
\begin{tabular}{clc}
\hline
\textbf{Features}                                                                & \textbf{ } & \textbf{Description}                                                                                \\ \hline
\textbf{\begin{tabular}[c]{@{}c@{}c@{}} 
Average nucleus area (ANA)\end{tabular}}    &                  & \textit{\begin{tabular}[c]{@{}c@{}}the average nucleus area of each region \end{tabular}}                       \\ \hline
\textbf{\begin{tabular}[c]{@{}c@{}c@{}} 
Average cytoplasm area (ACA)\end{tabular}}  &                  & \textit{\begin{tabular}[c]{@{}c@{}}the region which represents the  \\ subtraction of  total nucleus area from \\ total area of each region\end{tabular}}                       \\ \hline
\textbf{\begin{tabular}[c]{@{}c@{}c@{}} Nucleus-cytoplasm ratio (NCR)\end{tabular}} &                  & \textit{\begin{tabular}[c]{@{}c@{}}the division result of total nucleus area \\ to  the total the cytoplasm area \\ in each region\end{tabular}} \\ \hline
\textbf{\begin{tabular}[c]{@{}c@{}c@{}} 
Nucleus perimeter (NP)\end{tabular}}        &                  & \textit{\begin{tabular}[c]{@{}c@{}}the length of the line which surrounds \\ the nucleus  in each region\end{tabular}}                       \\ \hline
\textbf{\begin{tabular}[c]{@{}c@{}c@{}} 
Border irregularity (BI)\end{tabular}}      &                  & \textit{\begin{tabular}[c]{@{}c@{}}obtained by dividing length  of the \\   uniform ellipses that fit the nucleus \\ to the circumference of the related nucleus\end{tabular}}                       \\ \hline
\textbf{\begin{tabular}[c]{@{}c@{}c@{}}  Hyperkromosis index (HI)\end{tabular}}      &                  & \textit{\begin{tabular}[c]{@{}c@{}}represents the standard deviation value of   \\ the parabasal cells with respect \\ to the cells of the same lesion\end{tabular}}                       \\ \hline
\textbf{\begin{tabular}[c]{@{}c@{}}
Polarity lossindex (PI)\end{tabular}}       &                  & \textit{\begin{tabular}[c]{@{}c@{}}The average angle between the basal  \\ membrane and the  major axis \\ of all nucleus\end{tabular}}                       \\ \hline
\end{tabular}
\end{table}


The data set of feature vectors is imbalanced. Different classifiers have been proposed in the literature for imbalanced data sets \cite{imbalanced_class_1,imbalanced_class_2,imbalanced_class_3}. The Weighted k-Nearest Neigbour (w-kNN) algorithm is one of these. In this study, w-kNN algorithm is preferred because of its fast operation and practical use. It is also another important criterion for selecting that it gives successful results for imbalanced data sets \cite{wknn1,wknn2}. The w-knn algorithm looks at the \textit{k} closest neighbors class as the k-NN algorithm. In addition, for each neighbor, the weight \textit{w} defined in $ \textit{w} = \frac{1}{d(\textbf{x}_i,\textbf{x}_q)^2}$ is assigned to classify according to the weight of the classes.  $d(.,.)$ is Euclidean distance function. If neighbour sample $\textbf{x}_i$ is far from query sample $\textbf{x}_q$, the effect on the classification is weak, and vice versa. 

\section{Results}
\label{sec:results} 

The similarities and differences in the diagnoses  of SEP image sections in the data set given by the two pathologists with respect to CIN-based grading. Diagonal values refer to the number of images which have the same diagnosis of two pathologist. The agreement ratio of the pathologists is 73\% in the classification of the SEP images with respect to CIN-based grading are shown in Table \ref{tab:Pathologst1VSPathologist2TripleSystem}.

\begin{table}[!ht]
\centering
\caption{Agreement/disagreement of the pathologists experts in diagnosis of SEP with respect to CIN-based grading.}
\label{tab:Pathologst1VSPathologist2TripleSystem}
\begin{tabular}{|c|l|l|l|l|l|l|}
\hline
\multicolumn{1}{|l|}{}     & \multicolumn{6}{c|}{Pathologist 2}               \\ \hline
\multirow{6}{*}{Pathologist 1} &        & Normal & CIN1 & CIN2 & CIN3 & Total \\ \cline{2-7} 
                           & Normal & 354    & 37   & 0    & 0    & 391   \\ \cline{2-7} 
                           & CIN1   & 88     & 156  & 8    & 0    & 252   \\ \cline{2-7} 
                           & CIN2   & 13     & 52   & 71  & 15   & 151   \\ \cline{2-7} 
                           & CIN3   & 4      & 9   & 22   & 128  & 163   \\ \cline{2-7} 
                           & Total  & 459    & 254  & 101  & 143  & 957   \\ \hline
\end{tabular}
\end{table}

The agreement/disagreement of pathologist 1 and pathologist 2 to the final diagnosis of each SEP image patch are presented in Table \ref{tab:P1P2VSFD_CIN}. Final diagnosis is determined according to the disagreement of pathologists for a SEP image patch. The SEP that are not labeled as the same by the pathologists are then observed from the same tissue stained with p16 and Ki67 immunhistochemical dyes.

\begin{table}[!ht]
\centering
\caption{Agreement/disagreement of the pathologist 1, pathologist 2 and the proposed method with final diagnosis in diagnosing of SEP with respect to CIN-based grading.}
\label{tab:P1P2VSFD_CIN}
\begin{tabular}{cc|cccc||cccc||cccc}
\hline \hline
\multicolumn{2}{c}{\multirow{2}{*}{\textbf{CIN Grading}}} & \multicolumn{4}{c}{\textbf{P1}} & \multicolumn{4}{c}{\textbf{P2}} & \multicolumn{4}{c}{\textbf{Proposed}}\\ \cline{3-14} 
\multicolumn{2}{c}{}                             & \textbf{N}    & \textbf{C1}  & \textbf{C2}  & \textbf{C3}  & \textbf{N}    & \textbf{C1}  & \textbf{C2}  & \textbf{C3}  & \textbf{N}    & \textbf{C1}  & \textbf{C2}  & \textbf{C3} \\ \hline 
\multirow{4}{*}{\textbf{Final Diagnosis }}     & \textbf{N}         &  \textbf{377}   & 74  & 18  & 2   &  \textbf{424}  & 44  & 3   & 0  &  \textbf{386}   & 61  & 17  & 7  \\ 
                                              & \textbf{C1}        & 13   &  \textbf{176} & 37  & 14  & 26   &  \textbf{193} & 16  & 5 & 116   &  \textbf{91} & 19  & 14  \\
                                              & \textbf{C2}        & 0    & 2   &  \textbf{87}  & 18  & 8    & 17  &  \textbf{79}  & 3 & 13    & 30  &  \textbf{47}  & 17  \\  
                                              & \textbf{C3}        & 1    & 0   & 9   &  \textbf{129} & 1    & 0   & 3   &  \textbf{135} & 4    & 15   & 18   &  \textbf{102} \\  
                                              &  \textbf{Tot.}    & 391  & 252 & 151 & 163 & 459  & 254 & 101 & 143 & 519  & 197 & 101 & 140 \\ \hline \hline
\end{tabular}
\end{table}

Table \ref{tab:P1P2VSFD_CIN} represents the agreement between the final diagnosis and two pathologists with respect to CIN-based grading system. It can be observed that pathologist 2 has more compatible diagnosis result than pathologist 1 with final diagnosis. However, pathologist 1 has more consistent diagnosis in CIN2 SEP image patches. An important information to be drawn from the table is that the number of windowed classes (labelling CIN1 instead of Normal tissue; Normal or CIN2 instead of CIN1; CIN1 or CIN3 instead of CIN2 and CIN2 instead of CIN3) is high over. 

Another system that pathologists pay attention to while diagnosing tissues is the SIL-based grading system. In this system, the CIN2 grade is assumed to be two level, CIN3-like and CIN1-like. CIN2 lesion which resembles CIN3 and CIN3 are expressed as HSIL; CIN2 which resembles CIN1 and CIN1 are expressed as LSIL. The treatment of precursor lesions of cervical cancer varies according to LSIL and HSIL. Table \ref{tab:P1P2VSFD_SIL} represents the agreement between the final diagnosis and two pathologists with respect to SIL-based grading system. Pathologist 1 has more accurate results than pathologist 2 in normal and LSIL while pathologist 2 has more accurate result in diagnosing HSIL. If the diagnosis agreement of the pathologists according to the Table \ref{tab:P1P2VSFD_CIN} and Table \ref{tab:P1P2VSFD_SIL} are compared, it can be observed that pathologists makes more consistent diagnosis in SIL-based grading system. 

\begin{table}[!ht]
\centering
\caption{Agreement/disagreement of the Pathologists with final diagnosis in the diagnosis of SEP with respect to SIL-based grading.}
\label{tab:P1P2VSFD_SIL}
\begin{tabular}{cc|ccc||ccc}
\hline \hline
\multicolumn{2}{c}{\multirow{2}{*}{\textbf{SIL Grading}}} & \multicolumn{3}{c}{\textbf{P1}} & \multicolumn{3}{c}{\textbf{P2}} \\ \cline{3-8} 
\multicolumn{2}{c}{}                          					   & \textbf{N}    	   & \textbf{LSIL}  & \textbf{HSIL}  & \textbf{N}    & \textbf{LSIL}  & \textbf{HSIL}   \\ \hline 
\multirow{3}{*}{\textbf{Final Diagnosis}}     & \textbf{N}         &  \textbf{377}   & 74  & 20   &  \textbf{424}  & 44  & 3     \\ 
                                              & \textbf{LSIL}        & 13   &  \textbf{176} & 51    & 26   &  \textbf{193} & 21     \\  
                                              & \textbf{HSIL}        & 1    & 2   &  \textbf{243}  & 9    & 17  &  \textbf{220}   \\   
                                              & \textbf{Total}     & 391  & 252 & 314  & 459  & 254 & 244 \\ \hline \hline
\end{tabular}
\end{table}

Table \ref{tab:PMDTVSFD_CIN} represents the agreement between the final diagnosis with the proposed method  and the DT method which is the one of the best known algorithms used in diagnosis of cervical cancer grading depending on CIN-based grading system. 

\begin{table}[!ht]
\centering
\caption{Agreement/disagreement of the proposed method and Delaunay Triangulation(DT) with final diagnosis in the diagnosis of SEP with respect to CIN-based grading.}
\label{tab:PMDTVSFD_CIN}
\begin{tabular}{cc|cccc||cccc}
\hline \hline
\multicolumn{2}{c}{\multirow{2}{*}{\textbf{CIN Grading}}} & \multicolumn{4}{c}{\textbf{Proposed}} & \multicolumn{4}{c}{\textbf{DT}} \\ \cline{3-10} 
\multicolumn{2}{c}{}                             & \textbf{N}    & \textbf{C1}  & \textbf{C2}  & \textbf{C3}  & \textbf{N}    & \textbf{C1}  & \textbf{C2}  & \textbf{C3}  \\ \hline 
\multirow{4}{*}{\textbf{Final Diagnosis}}     & \textbf{N}         & \textbf{386}   & 61  & 17  & 7   & \textbf{471}  & 0  & 0   & 0   \\ 
                                              & \textbf{C1}        & 116   & \textbf{91} & 19  & 14  & 240   & \textbf{0} & 0  & 0   \\  
                                              & \textbf{C2}        & 13    & 30   & \textbf{47}  & 17  & 107    & 0  & \textbf{0}  & 0   \\  
                                              & \textbf{C3}        & 4    & 15   & 18   & \textbf{102} & 139    & 0   & 0   & \textbf{0} \\  
                                              & \textbf{Total}     & 519  & 197 & 101 & 140 & 0  & 0 & 0 & 0 \\ \hline \hline
\end{tabular}
\end{table}

Table \ref{tab:PMDTVSFD_CIN} represents the agreement between the final diagnosis and the proposed method with respect to CIN-based grading system. Normal and CIN3 SEP patches prediction are classified accurately. However, predicting CIN1 and CIN2 SEP patches is lower while comparing with CIN1 and CIN3. The classification accuracy of the proposed method is approximately $65.4\%$. If the results obtained from the proposed method is compared with the pathologists, it can be said that the CAD system developed in this study should be improved in order to be used as a secondary decision system to help the pathologist in predicting cervical cancer precursor lesion grade.


Table \ref{tab:PMDTVSFD_SIL} represents the agreement between the final diagnosis and the proposed method with respect to SIL-based grading system. The classification performance of the proposed study is again less accurate than the pathologist. However, the results obtained from the SIL-based grading system of the proposed method is improved to $70.5\%$ comparing to the CIN-based grading system.

\begin{table}[!ht]
\centering
\caption{Agreement/disagreement of the proposed and DT method with final diagnosis in diagnosis of SEP with respect to SIL-based grading.}
\label{tab:PMDTVSFD_SIL}
\begin{tabular}{cc|ccc}
\hline \hline
\multicolumn{2}{c}{\multirow{2}{*}{\textbf{SIL Grading}}} & \multicolumn{3}{c}{\textbf{Proposed}} \\ \cline{3-5} 
\multicolumn{2}{c}{}                    & \textbf{N}      & \textbf{LSIL}  & \textbf{HSIL}   \\ \hline 
\multirow{3}{*}{\textbf{Final Diagnosis}} & \textbf{N}    &  \textbf{381}   & 57  & 33  \\ 
                                        & \textbf{LSIL} & 106   &  \textbf{91} & 43    \\  
                                              & \textbf{HSIL}        & 16    & 27   &  \textbf{203}  \\   
                                              & \textbf{Total}     & 503  & 175 & 279  \\ \hline \hline
\end{tabular}
\end{table}

\section{Discussion} 
\label{sec:discussion} 

In this study, it is aimed to translate the evaluations of pathologists which have subjectivity on cervical dysplasia  to into numerical values and to develop a `Computer Assisted Diagnostic Auxiliary Systems (CADAS)' as a result. Our study on cervical dysplasia has the largest data set according to the similar studies available in the literature. Furthermore, the fact that the diagnoses are given by two pathologists, and the reassessment and determination of the definitive diagnosis during inconsistent cases increased the reliability of the CADAS training set. The developed CADAS promises to be used as an assistant system in the future because of numerical values that are found to be in parallel with the diagnostic parameters used by the pathologists (such as ratio of nucleus to cytoplasm, nucleus boundary irregularity, polarity loss and hyperchromaticity) and statistically significant. The studies in the literature are mostly designed by engineers and the contribution of pathologists is very limited. For this reason, there are some shortcomings when viewed from the perspective of pathology and clinical approach. In the development of a CADAS to be used in pathology, the presence of pathologists at every step is a necessary requirement.

\section{Conclusion} 
\label{sec:conclusion} 
In this paper, we present a new benchmark data set of cervical cancer precursor lesions, which we make available to the scientific community for grading the cervical intraepithelial neoplasia. Each image in the data set is labeled by two pathologist experts to reveal the inter-observer variability. In case of different diagnoses, p16 and Ki67 immunohistochemical dyes are used to decide a final diagnosis (ground truth). There are also papilla areas that seriously affect the performance of automated methods which makes this study unique as far as we know. A morphological analysis based feature extraction method is also proposed in the study for the grading of cervical cancer precursor lesions. The result of the study is also compared with each pathologist expert and the ground truth. The results show that CAD systems could be used as a secondary decision system for experts with some improvement. It is aimed to improve the classification performance of our CAD system by developing up-to-date image processing and machine learning algorithms especially types of deep learning.

\section{Data Availability} 
\label{sec:Dataavailability} 
The materials and datasets generated during and/or analysed during the current study are available from the \href{http://simplab.yildiz.edu.tr/?q=sources}{http://simplab.yildiz. edu.tr/?q=sources} on reasonable request.

\section*{Conflict of Interest} 
All authors declare that they have no conflict of interest.

\section*{Competing Interests} 
The authors declare no competing interests.
 
\section*{Acknowledgement}
\noindent
Authors also would like to thank  Argenit Company and Istanbul Medipol University Hospital for providing and annotating the whole slide histopathological images of cervical cancer precursor lesions  image data set. The authors state no conflict of interest and have nothing to disclose.

\bibliographystyle{naturemag-doi}
\bibliography{references}

\begin{thebibliography}{10}
\urlstyle{rm}
\expandafter\ifx\csname url\endcsname\relax
  \def\url#1{\texttt{#1}}\fi
\expandafter\ifx\csname urlprefix\endcsname\relax\def\urlprefix{URL }\fi
\expandafter\ifx\csname doiprefix\endcsname\relax\def\doiprefix{DOI: }\fi
\providecommand{\bibinfo}[2]{#2}
\providecommand{\eprint}[2][]{\url{#2}}

\bibitem{torre2015global}
\bibinfo{author}{Torre, L.~A.} \emph{et~al.}
\newblock \bibinfo{journal}{\bibinfo{title}{Global cancer statistics, 2012}}.
\newblock {\emph{\JournalTitle{CA Cancer J. Clin.}}}
  \textbf{\bibinfo{volume}{65}}, \bibinfo{pages}{87--108}
  (\bibinfo{year}{2015}).

\bibitem{stoler2000human}
\bibinfo{author}{Stoler, M.~H.}
\newblock \bibinfo{journal}{\bibinfo{title}{Human papillomaviruses and cervical
  neoplasia: a model for carcinogenesis}}.
\newblock {\emph{\JournalTitle{Int. J. Gynecol. Pathol.}}}
  \textbf{\bibinfo{volume}{19}}, \bibinfo{pages}{16--28}
  (\bibinfo{year}{2000}).

\bibitem{van2018hpv}
\bibinfo{author}{Van~Zummeren, M.} \emph{et~al.}
\newblock \bibinfo{journal}{\bibinfo{title}{Hpv e4 expression and dna
  hypermethylation of cadm1, mal, and mir124-2 genes in cervical cancer and
  precursor lesions}}.
\newblock {\emph{\JournalTitle{Modern Pathology}}} \bibinfo{pages}{1}
  (\bibinfo{year}{2018}).

\bibitem{zur2009papillomaviruses}
\bibinfo{author}{Zur~H., H.}
\newblock \bibinfo{journal}{\bibinfo{title}{Papillomaviruses in the causation
  of human cancers—a brief historical account}}.
\newblock {\emph{\JournalTitle{Virol. J.}}} \textbf{\bibinfo{volume}{384}},
  \bibinfo{pages}{260--265} (\bibinfo{year}{2009}).

\bibitem{stoler2000advances}
\bibinfo{author}{Stoler, M.~H.}
\newblock \bibinfo{journal}{\bibinfo{title}{Advances in cervical screening
  technology}}.
\newblock {\emph{\JournalTitle{Mod. Pathol.}}} \textbf{\bibinfo{volume}{13}},
  \bibinfo{pages}{275--284} (\bibinfo{year}{2000}).

\bibitem{cox2013historical}
\bibinfo{author}{Cox, J.~T.}, \bibinfo{author}{Wilkinson, E.~J.} \&
  \bibinfo{author}{O’connor, D.~M.}
\newblock \bibinfo{journal}{\bibinfo{title}{Historical perspective: terminology
  for lower anogenital tract pathology}}.
\newblock {\emph{\JournalTitle{AJSP: Reviews \& Reports}}}
  \textbf{\bibinfo{volume}{18}}, \bibinfo{pages}{158--167}
  (\bibinfo{year}{2013}).

\bibitem{darragh2012lower}
\bibinfo{author}{Darragh, T.~M.} \emph{et~al.}
\newblock \bibinfo{journal}{\bibinfo{title}{The lower anogenital squamous
  terminology standardization project for hpv-associated lesions: background
  and consensus recommendations from the college of american pathologists and
  the american society for colposcopy and cervical pathology}}.
\newblock {\emph{\JournalTitle{Arch. Path. Lab. Med.}}}
  \textbf{\bibinfo{volume}{136}}, \bibinfo{pages}{1266--1297}
  (\bibinfo{year}{2012}).

\bibitem{mitra2015cervical}
\bibinfo{author}{Mitra, A.} \emph{et~al.}
\newblock \bibinfo{journal}{\bibinfo{title}{Cervical intraepithelial neoplasia
  disease progression is associated with increased vaginal microbiome
  diversity}}.
\newblock {\emph{\JournalTitle{Scientific reports}}}
  \textbf{\bibinfo{volume}{5}}, \bibinfo{pages}{16865} (\bibinfo{year}{2015}).

\bibitem{stoler2001interobserver}
\bibinfo{author}{Stoler, S.~M., M.~H.} \emph{et~al.}
\newblock \bibinfo{journal}{\bibinfo{title}{Interobserver reproducibility of
  cervical cytologic and histologic interpretations: realistic estimates from
  the ascus-lsil triage study}}.
\newblock {\emph{\JournalTitle{Jama}}} \textbf{\bibinfo{volume}{285}},
  \bibinfo{pages}{1500--1505} (\bibinfo{year}{2001}).

\bibitem{mccluggage1996interobserver}
\bibinfo{author}{McCluggage, W.} \emph{et~al.}
\newblock \bibinfo{journal}{\bibinfo{title}{Interobserver variation in the
  reporting of cervical colposcopic biopsy specimens: Comparison of grading
  systems}}.
\newblock {\emph{\JournalTitle{J. Clin. Path.}}} \textbf{\bibinfo{volume}{49}},
  \bibinfo{pages}{833--835} (\bibinfo{year}{1996}).

\bibitem{mccluggage1998inter}
\bibinfo{author}{McCluggage, W.} \emph{et~al.}
\newblock \bibinfo{journal}{\bibinfo{title}{Inter-and intra-observer variation
  in the histopathological reporting of cervical squamous in traepithelial
  lesion susing a modified bethesda grading system}}.
\newblock {\emph{\JournalTitle{Br. J. Obstet. Gynaecol.}}}
  \textbf{\bibinfo{volume}{105}}, \bibinfo{pages}{206--210}
  (\bibinfo{year}{1998}).

\bibitem{doorbar2007papillomavirus}
\bibinfo{author}{Doorbar, J.}
\newblock \bibinfo{journal}{\bibinfo{title}{Papillomavirus life cycle
  organization and biomarker selection}}.
\newblock {\emph{\JournalTitle{Dis. Markers}}} \textbf{\bibinfo{volume}{23}},
  \bibinfo{pages}{297--313} (\bibinfo{year}{2007}).

\bibitem{al2012digital}
\bibinfo{author}{Al-Janabi, S.}, \bibinfo{author}{Huisman, A.} \&
  \bibinfo{author}{J., V. D.~P.}
\newblock \bibinfo{journal}{\bibinfo{title}{Digital pathology: current status
  and future perspectives}}.
\newblock {\emph{\JournalTitle{Histopathology}}} \textbf{\bibinfo{volume}{61}},
  \bibinfo{pages}{1--9} (\bibinfo{year}{2012}).

\bibitem{madabhushi2016image}
\bibinfo{author}{Madabhushi, A.} \& \bibinfo{author}{Lee, G.}
\newblock \bibinfo{journal}{\bibinfo{title}{Image analysis and machine learning
  in digital pathology: Challenges and opportunities}}.
\newblock {\emph{\JournalTitle{Med. Image Anal.}}}
  \textbf{\bibinfo{volume}{33}}, \bibinfo{pages}{170 -- 175}
  (\bibinfo{year}{2016}).

\bibitem{he2012histology}
\bibinfo{author}{He, L.}, \bibinfo{author}{Long, L.~R.},
  \bibinfo{author}{Antani, S.} \& \bibinfo{author}{Thoma, G.~R.}
\newblock \bibinfo{journal}{\bibinfo{title}{Histology image analysis for
  carcinoma detection and grading}}.
\newblock {\emph{\JournalTitle{Comput. Methods Programs Biomed.}}}
  \textbf{\bibinfo{volume}{107}}, \bibinfo{pages}{538--556}
  (\bibinfo{year}{2012}).

\bibitem{de2013fusion}
\bibinfo{author}{De, S.} \emph{et~al.}
\newblock \bibinfo{journal}{\bibinfo{title}{A fusion-based approach for uterine
  cervical cancer histology image classification}}.
\newblock {\emph{\JournalTitle{Comput. Med. Imaging. Graph.}}}
  \textbf{\bibinfo{volume}{37}}, \bibinfo{pages}{475--487}
  (\bibinfo{year}{2013}).

\bibitem{guo2016nuclei}
\bibinfo{author}{Guo, P.} \emph{et~al.}
\newblock \bibinfo{journal}{\bibinfo{title}{Nuclei-based features for uterine
  cervical cancer histology image analysis with fusion-based classification}}.
\newblock {\emph{\JournalTitle{IEEE J. Biomed. Health Inform.}}}
  \textbf{\bibinfo{volume}{20}}, \bibinfo{pages}{1595--1607}
  (\bibinfo{year}{2016}).

\bibitem{naghdy2012computer}
\bibinfo{author}{Naghdy, G.}, \bibinfo{author}{Ros, M.~B.},
  \bibinfo{author}{Todd, C.} \emph{et~al.}
\newblock \bibinfo{title}{Computer aided decision support system for cervical
  cancer classification}.
\newblock In \emph{\bibinfo{booktitle}{Applications of Digital Image Processing
  XXXV}}, vol. \bibinfo{volume}{8499}, \bibinfo{pages}{849919}
  (\bibinfo{organization}{SPIE}, \bibinfo{year}{2012}).

\bibitem{wang2009assisted}
\bibinfo{author}{Wang, D., Y.and~Crookes}, \bibinfo{author}{Eldin, O.~S.},
  \bibinfo{author}{Wang, P., S.and~Hamilton} \& \bibinfo{author}{Diamond, J.}
\newblock \bibinfo{journal}{\bibinfo{title}{Assisted diagnosis of cervical
  intraepithelial neoplasia (cin)}}.
\newblock {\emph{\JournalTitle{IEEE J. Sel. Top. Signal. Process.}}}
  \textbf{\bibinfo{volume}{3}}, \bibinfo{pages}{112--121}
  (\bibinfo{year}{2009}).

\bibitem{keenan2000automated}
\bibinfo{author}{Keenan, S.} \emph{et~al.}
\newblock \bibinfo{journal}{\bibinfo{title}{An automated machine vision system
  for the histological grading of cervical intraepithelial neoplasia (cin)}}.
\newblock {\emph{\JournalTitle{J. Pathol.}}} \textbf{\bibinfo{volume}{192}},
  \bibinfo{pages}{351--362} (\bibinfo{year}{2000}).

\bibitem{melo2016expression}
\bibinfo{author}{de~Melo, F.}, \bibinfo{author}{Lancellotti, C.} \&
  \bibinfo{author}{da~Silva, M.}
\newblock \bibinfo{journal}{\bibinfo{title}{Expression of the
  immunohistochemical markers p16 and ki-67 and their usefulness in the
  diagnosis of cervical intraepithelial neoplasms}}.
\newblock {\emph{\JournalTitle{Rev. Bras. Ginicol. Obstet.}}}
  \textbf{\bibinfo{volume}{38}}, \bibinfo{pages}{82--87}
  (\bibinfo{year}{2016}).

\bibitem{galgano2010using}
\bibinfo{author}{Galgano, M.~T.} \emph{et~al.}
\newblock \bibinfo{journal}{\bibinfo{title}{Using biomarkers as objective
  standards in the diagnosis of cervical biopsies}}.
\newblock {\emph{\JournalTitle{Am. J. Surg. Path.}}}
  \textbf{\bibinfo{volume}{34}}, \bibinfo{pages}{1077} (\bibinfo{year}{2010}).

\bibitem{guo2011efficacy}
\bibinfo{author}{Guo, M.} \emph{et~al.}
\newblock \bibinfo{journal}{\bibinfo{title}{Efficacy of p16 and proexc
  immunostaining in the detection of high-grade cervical intraepithelial
  neoplasia and cervical carcinoma}}.
\newblock {\emph{\JournalTitle{Am. J. Clin. Pathol.}}}
  \textbf{\bibinfo{volume}{135}}, \bibinfo{pages}{212--220}
  (\bibinfo{year}{2011}).

\bibitem{lim2016efficacy}
\bibinfo{author}{Lim, S.}, \bibinfo{author}{Lee, M.}, \bibinfo{author}{Cho,
  I.}, \bibinfo{author}{Hong, R.} \& \bibinfo{author}{Lim, S.}
\newblock \bibinfo{journal}{\bibinfo{title}{Efficacy of p16 and ki-67
  immunostaining in the detection of squamous intraepithelial lesions in a
  high-risk hpv group}}.
\newblock {\emph{\JournalTitle{Oncol. Lett.}}} \textbf{\bibinfo{volume}{11}},
  \bibinfo{pages}{1447--1452} (\bibinfo{year}{2016}).

\bibitem{ozaki2011biomarker}
\bibinfo{author}{Ozaki, S.}, \bibinfo{author}{Zen, Y.} \&
  \bibinfo{author}{Inoue, M.}
\newblock \bibinfo{journal}{\bibinfo{title}{Biomarker expression in cervical
  intraepithelial neoplasia: potential progression predictive factors for
  low-grade lesions}}.
\newblock {\emph{\JournalTitle{Hum. Pathol.}}} \textbf{\bibinfo{volume}{42}},
  \bibinfo{pages}{1007--1012} (\bibinfo{year}{2011}).

\bibitem{achanta2012slic}
\bibinfo{author}{Achanta, R.} \emph{et~al.}
\newblock \bibinfo{journal}{\bibinfo{title}{Slic superpixels compared to
  state-of-the-art superpixel methods}}.
\newblock {\emph{\JournalTitle{IEEE Trans. Pattern. Anal. Mach. Intell.}}}
  \textbf{\bibinfo{volume}{34}}, \bibinfo{pages}{2274--2282}
  (\bibinfo{year}{2012}).

\bibitem{najar2017comparison}
\bibinfo{author}{Najar, F.}, \bibinfo{author}{Bourouis, S.},
  \bibinfo{author}{Bouguila, N.} \& \bibinfo{author}{Belghith, S.}
\newblock \bibinfo{title}{A comparison between different gaussian-based mixture
  models}.
\newblock In \emph{\bibinfo{booktitle}{2017 IEEE/ACS 14th International
  Conference on Computer Systems and Applications, AICCSA'17}},
  \bibinfo{pages}{704--708} (\bibinfo{organization}{IEEE},
  \bibinfo{year}{2017}).

\bibitem{imbalanced_class_1}
\bibinfo{author}{Mazurowski, M.~A.} \emph{et~al.}
\newblock \bibinfo{journal}{\bibinfo{title}{Training neural network classifiers
  for medical decision making: The effects of imbalanced datasets on
  classification performance}}.
\newblock {\emph{\JournalTitle{Neural Netw.}}} \textbf{\bibinfo{volume}{21}},
  \bibinfo{pages}{427--436} (\bibinfo{year}{2008}).

\bibitem{imbalanced_class_2}
\bibinfo{author}{Hong, X.}, \bibinfo{author}{Chen, S.} \&
  \bibinfo{author}{Harris, C.~J.}
\newblock \bibinfo{journal}{\bibinfo{title}{A kernel-based two-class classifier
  for imbalanced data sets}}.
\newblock {\emph{\JournalTitle{IEEE Trans. Neural Netw.}}}
  \textbf{\bibinfo{volume}{18}}, \bibinfo{pages}{28--41}
  (\bibinfo{year}{2007}).

\bibitem{imbalanced_class_3}
\bibinfo{author}{Zhuang, L.} \& \bibinfo{author}{Dai, H.}
\newblock \bibinfo{journal}{\bibinfo{title}{Parameter optimization of
  kernel-based one-class classifier on imbalance learning}}.
\newblock {\emph{\JournalTitle{J. Comput.}}} \textbf{\bibinfo{volume}{1}},
  \bibinfo{pages}{32--40} (\bibinfo{year}{2006}).

\bibitem{wknn1}
\bibinfo{author}{Liu, W.} \& \bibinfo{author}{Chawla, S.}
\newblock \bibinfo{title}{Class confidence weighted knn algorithms for
  imbalanced data sets}.
\newblock In \emph{\bibinfo{booktitle}{Pacific-Asia Conference on Knowledge
  Discovery and Data Mining}}, \bibinfo{pages}{345--356}
  (\bibinfo{organization}{Springer}, \bibinfo{year}{2011}).

\bibitem{wknn2}
\bibinfo{author}{Zuo, W.}, \bibinfo{author}{Lu, W.}, \bibinfo{author}{Wang, K.}
  \& \bibinfo{author}{Zhang, H.}
\newblock \bibinfo{title}{Diagnosis of cardiac arrhythmia using kernel
  difference weighted knn classifier}.
\newblock In \emph{\bibinfo{booktitle}{Computers in Cardiology}},
  \bibinfo{pages}{253--256} (\bibinfo{organization}{IEEE},
  \bibinfo{year}{2008}).

\end{thebibliography}

\section*{Author contributions statement}

A.A. , A.U. and N.C. designed and performed the experiments, G.B.,  L.D.A and  B.U.T were involved in planning and supervised the work,  A.C., I.T. and B.M. contributed to the design and implementation of the research. All authors reviewed the manuscript.

\end{document}